\RequirePackage{fix-cm}
\documentclass[twocolumn]{svjour3}          
\smartqed  
\usepackage{xcolor}
\usepackage{tabularx}
\usepackage{multirow}
\usepackage{graphicx}
\usepackage{makecell}
\usepackage{amsmath}
\usepackage{color}
\usepackage{float}
\usepackage{rotating}
\usepackage{placeins}
\usepackage{amsfonts}
\usepackage{bm}

\def\eg{\textit{e.g.}}
\def\ie{\textit{i.e.}}

\usepackage[round,authoryear,comma,sort&compress,numbers]{natbib} 
\usepackage[colorlinks=true,linkcolor=blue,citecolor=black,urlcolor=blue]{hyperref}

\usepackage{diagbox}
\usepackage{cancel} 
\usepackage{mathrsfs} 
\usepackage{eqnarray} 

\usepackage{ctable} 

\usepackage{algpseudocode}%
\usepackage{listings}%
\usepackage{bbding}
\usepackage{bbm}
\usepackage[ruled]{algorithm2e}
\usepackage[caption=false, font=footnotesize]{subfig}
\usepackage{fontawesome}
\usepackage{array}

\usepackage{colortbl}  
\usepackage{arydshln}
\usepackage{longtable}
\usepackage{wrapfig}
\usepackage{bbold}
\usepackage{caption}
\usepackage{pifont}

\newcommand{\PreserveBackslash}[1]{\let\temp=\\#1\let\\=\temp}
\newcolumntype{C}[1]{>{\PreserveBackslash\centering}p{#1}}
\newcolumntype{R}[1]{>{\PreserveBackslash\raggedleft}p{#1}}
\newcolumntype{L}[1]{>{\PreserveBackslash\raggedright}p{#1}}

\def\eg{\textit{e.g.}}
\def\ie{\textit{i.e.}}

\def\vs{\textit{vs.}}
\definecolor{battleshipgrey}{rgb}{0.52, 0.52, 0.51}
\definecolor{capri}{rgb}{0.0, 0.75, 1.0}
\definecolor{mediumspringgreen}{rgb}{0.0, 0.98, 0.6}

\journalname{International Journal of Computer Vision}
\begin{document}

\title{SeaFormer++: Squeeze-enhanced Axial Transformer for Mobile Visual Recognition
}


\author{
	Qiang Wan$^1$, \and
	Zilong Huang$^2$, \and
        Jiachen Lu$^1$, \and
        Gang Yu$^3$, \and
        Li Zhang$^1$\textsuperscript{\faEnvelopeO}
}




\institute{
	Corresponding author: Li Zhang  \at
             \email{lizhangfd@fudan.edu.cn}          \\
$^1$ School of Data Science, Fudan University, Shanghai, China \\
$^2$ ByteDance, Singapore \\
$^3$ Tencent, Shanghai, China \\
}
\date{6 Febuary 2025}

\maketitle

\begin{abstract}
Since the introduction of Vision Transformers, the landscape of many computer vision tasks (\eg, semantic segmentation), which has been overwhelmingly dominated by CNNs, recently has significantly revolutionized. 
However, the computational cost and memory requirement renders these methods unsuitable on the mobile device.
In this paper, we introduce a new method squeeze-enhanced Axial Transformer (SeaFormer) for mobile visual recognition.
Specifically, we design a generic attention block characterized by the formulation of squeeze Axial and detail enhancement.
It can be further used to create a family of backbone architectures with superior cost-effectiveness.
Coupled with a light segmentation head, we achieve the best trade-off between segmentation accuracy and latency on the ARM-based mobile devices on the ADE20K, Cityscapes Pascal Context and COCO-Stuff datasets.
Critically, we beat both the mobile-friendly rivals and Transformer-based counterparts with better performance and lower latency without bells and whistles.
Furthermore, we incorporate a feature upsampling-based multi-resolution distillation technique, further reducing the inference latency of the proposed framework. 
Beyond semantic segmentation, we further apply the proposed SeaFormer architecture to image classification and object detection problems, demonstrating the potential of serving as a versatile mobile-friendly backbone.
Our code and models are made publicly available at \url{https://github.com/fudan-zvg/SeaFormer}.
\end{abstract}
\section{Introduction}
As a fundamental problem in computer vision, semantic segmentation aims to assign a semantic class label to each pixel in an image.
Conventional methods rely on stacking local convolution kernel~\cite{long2015fully} to perceive the long-range structure information of the image.
Since the introduction of Vision Transformers~\cite{dosovitskiy2020image}, the landscape of semantic segmentation has significantly revolutionized. 
Transformer-based approaches~\cite{zheng2021rethinking, xie2021segformer} have remarkably demonstrated the capability of global context modeling.
However, the computational cost and memory requirement of Transformer render these methods unsuitable on mobile devices, especially for high-resolution imagery inputs.
Following conventional wisdom of efficient operation, local/window-based attention~\cite{luong2015effective,liu2021swin,huang2021shuffle,yuan2021hrformer}, Axial attention~\cite{huang2019ccnet,ho2019axial,wang2020axial}, dynamic graph message passing~\cite{zhang2020dynamic,zhang2023dynamic} and some lightweight attention mechanisms~\cite{hou2020strip,li2021towards,li2021global,li2020improving,liu2018generating,shen2021efficient,xu2021co,cao2019gcnet,woo2018cbam,wang2020linformer,choromanski2020rethinking,chen2017rethinking,mehta2021mobilevit} are introduced. 

\begin{figure*}
    \centering
    \includegraphics[width=\linewidth]{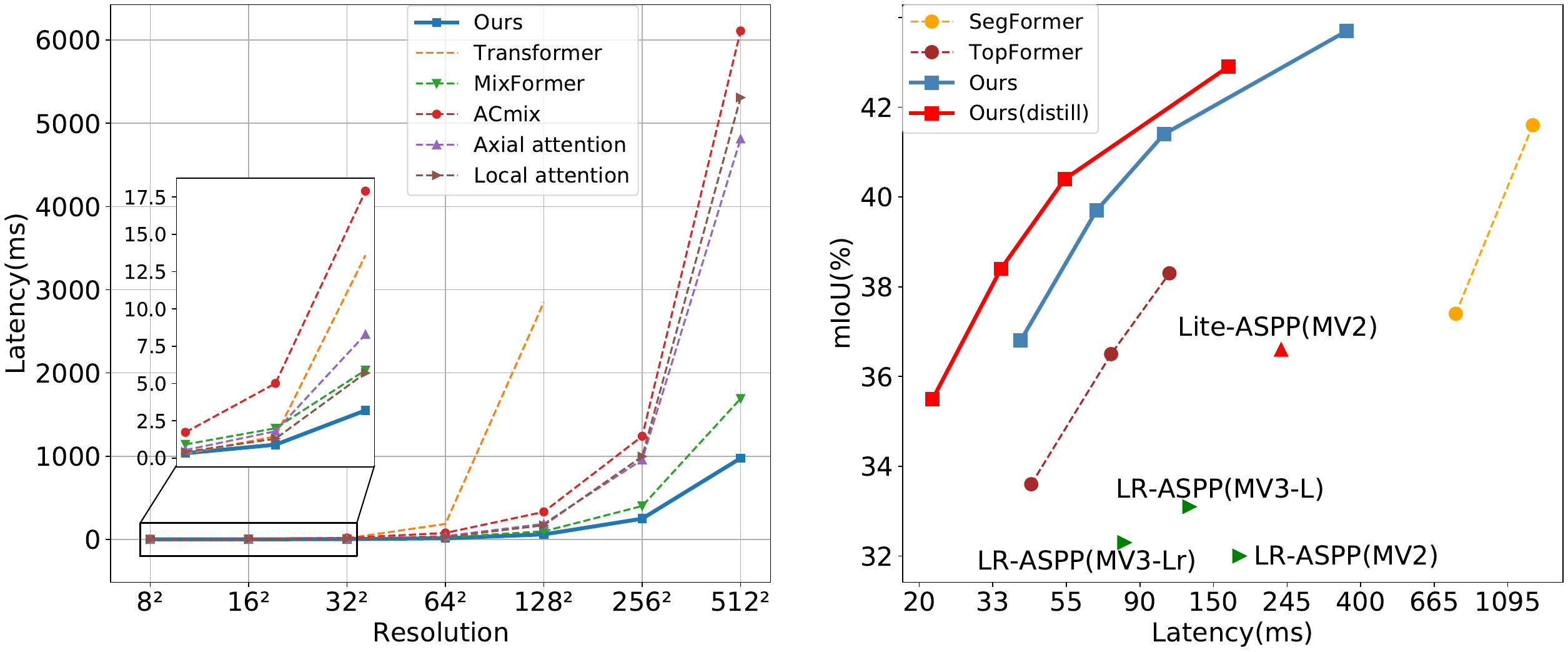}
    \caption{\textit{\textbf{Left}}: 
    Latency comparison with Transformer~\cite{vaswani2017attention}, MixFormer~\cite{chen2022mixformer}, ACmix~\cite{pan2022integration}, 
    Axial attention~\cite{ho2019axial} and 
    local attention~\cite{luong2015effective}. 
    It is measured with a single module of channel dimension 64 on a Qualcomm Snapdragon 865 processor. 
    \textit{\textbf{Right}}: 
    The mIoU versus latency on the ADE20K \textit{val} set. 
    MV2 means MobileNetV2~\cite{sandler2018mobilenetv2}. 
    MV3-L means MobileNetV3-Large~\cite{howard2019searching}. 
    MV3-Lr denotes MobileNetV3-Large-reduce~\cite{howard2019searching}. 
    The latency is measured on a single Qualcomm Snapdragon 865, and only an ARM CPU core is used for speed testing. No other means of acceleration, e.g., GPU or quantification, is used. For figure \textit{Right}, the input size is 512×512. SeaFormer achieves superior trade-off between mIoU and latency.}
    \label{fig:latency_comp}
\end{figure*}
However, these advances are still insufficient to satisfy the design requirements and constraints for mobile devices due to the high latency on the high-resolution inputs (see Figure~\ref{fig:latency_comp}). 
Recently there is a surge of interest in building a Transformer-based semantic segmentation.
In order to reduce the computation cost at high resolution, 
TopFormer~\cite{zhang2022topformer} dedicates to applying the global attention at a $1/64$ scale of the original input, which definitely harms the segmentation performance.
To solve the dilemma of high-resolution computation for pixel-wise segmentation task and low latency requirements on the mobile device in a performance-harmless way, we propose a family of mobile-friendly Transformer-based semantic segmentation models, dubbed squeeze-enhanced Axial Transformer (SeaFormer), which reduces the computational complexity of axial attention from $\mathcal{O}((H+W)HW)$ to $\mathcal{O}(HW)$, to achieve superior accuracy-efficiency trade-off on mobile devices and fill the vacancy of mobile-friendly efficient Transformer.

The core building block \textit{squeeze-enhanced Axial attention} (SEA attention) seeks to squeeze (pool) the input feature maps along the horizontal/vertical axis into a compact column/row and computes self-attention.
We concatenate query, keys and values to compensate the detail information sacrificed during squeeze and then feed it into a depth-wise convolution layer to enhance local details. The SEA attention module is a novel enhancement designed to reduce computational complexity, making it suitable for mobile devices without sacrificing performance. 

Coupled with a light segmentation head and several lightweight fusion blocks, our design (see Figure~\ref{fig:pipeline}) with the proposed SeaFormer layer in the small-scale feature is capable of conducting high-resolution image semantic segmentation with low latency on the mobile device. Our dual-branch network architecture incorporates these unique design elements that optimize efficiency and accuracy for mobile applications.
As shown in Figure~\ref{fig:latency_comp}, the proposed SeaFormer outperforms other efficient neural networks on the ADE20K dataset with lower latency. 
In particular, SeaFormer-Base is superior to the lightweight CNN counterpart MobileNetV3 (41.0 \vs 33.1 mIoU) with lower latency (106ms \vs 126ms) on an ARM-based mobile device. 
Furthermore, we have proposed a multi-resolution distillation framework that further enhances the model's efficiency and accuracy.

We make the following \textbf{contributions}:
\textbf{(i)}
We introduce a novel squeeze-enhanced Axial Transformer (SeaFormer) framework for mobile semantic segmentation;
\textbf{(ii)}
Critically, we design a generic attention block characterized by the formulation of squeeze Axial and detail enhancement;
It can be used to create a family of backbone architectures with superior cost-effectiveness;
\textbf{(iii)}
We show top performance on the ADE20K and Cityscapes datasets, beating both the mobile-friendly rival and Transformer-based segmentation model with clear margins;
\textbf{(iv)}
Beyond semantic segmentation, we further apply the proposed SeaFormer architecture to 
the image classification problem, demonstrating the potential of serving as a versatile mobile-friendly backbone.

A preliminary version of this work was presented in ICLR 2023~\cite{wan2023seaformer}.
In this paper, we have further extended our conference version as follows:
{\bf (i)} 
We propose an adaptive squeeze and expand method in Squeeze-axial attention, using a learnable mask to map all tokens of query/key/values to a single token in each row and column;
{\bf (ii)} 
We present a feature upsampling based multi-resolution distillation approach, further reducing the inference latency of the proposed framework.
\section{Related work}
\label{Rel_wk}

\subsection{Combination of Transformers and convolution}
Convolution is relatively efficient but not suitable to capture long-range dependencies and vision Transformer has the powerful capability with a global receptive field but lacks efficiency due to the computation of self attention. 
In order to make full use of both of their advantages, MobileViT \cite{mehta2021mobilevit}, TopFormer \cite{zhang2022topformer}, LVT \cite{yang2022lite}, Mobile-Former \cite{chen2022mobile}, EdgeViTs \cite{pan2022edgevits}, MobileViTv2 \cite{mehta2022separable}, EdgeFormer \cite{zhang2022edgeformer} and EfficientFormer \cite{li2022efficientformer} are constructed as efficient ViTs by combining convolution with Transformers. MobileViT, Mobile-Former, TopFormer and EfficientFormer are restricted by Transformer blocks and have to trade off between efficiency and performance in model design. LVT, MobileViTv2 and EdgeViTs keep the model size small at the cost of relatively high computation, which also means high latency.

\subsection{Axial attention and variants}
Axial attention~\cite{huang2019ccnet, ho2019axial, wang2020axial} is designed to reduce the computational complexity of original global self-attention~\cite{vaswani2017attention}. 
It computes self-attention over a single axis at a time and stacks a horizontal and a vertical axial attention module to obtain the global receptive field. 
Strip pooling~\cite{hou2020strip} and Coordinate attention~\cite{hou2021coordinate} uses a band shape pooling window to pool along either the horizontal or the vertical dimension to gather long-range context. Kronecker Attention Networks~\cite{gao2020kronecker} uses the juxtaposition of horizontal and lateral average matrices to average the input matrices and performs attention operation. These methods and other similar implementations provide performance gains partly at considerably low computational cost compared with Axial attention. However, they ignore the lack of local details brought by the pooling/average operation. 

\subsection{Mobile semantic segmentation}
The mainstream of efficient segmentation methods are based on lightweight CNNs. 
DFANet~\cite{li2019dfanet} adopts a lightweight backbone to reduce computation cost and adds a feature aggregation module to refine high-level and low-level features. 
ICNet~\cite{zhao2018icnet} designs an image cascade network to speed up the algorithm, while BiSeNet~\cite{yu2018bisenet, yu2021bisenet} proposes two-stream paths for low-level details and high-level context information, separately. 
Fast-SCNN~\cite{poudel2019fast} shares the computational cost of the multi-branch network to yield a run-time fast segmentation CNN. 
TopFormer~\cite{zhang2022topformer} presents a new architecture with a combination of CNNs and ViT and achieves a good trade-off between accuracy and computational cost for mobile semantic segmentation. 
However, it is still restricted by the heavy computation load of global self-attention. 

\subsection{Knowledge distillation}
Knowledge Distillation (KD) is a widely used technique in model compression, where a smaller model (student) is trained to mimic the behavior of a larger, well performing model (teacher). The concept was first introduced by Hinton et al.~\cite{hinton2015distilling}, where the student model learns from the teacher's softened output probabilities. Since then, various distillation methods have been proposed~\cite{romero2014fitnets,park2019relational,tian2019contrastive,liu2022transkd,he2019knowledge,liu2019structured,yang2022cross,cho2019efficacy,heo2019comprehensive,kim2018paraphrasing,li2023curriculum,li2021online,mirzadeh2020improved,yang2019snapshot,yim2017gift,zhang2018deep,zhao2022decoupled,wang2024crosskd} to improve the learning process, focusing on feature-based distillation~\cite{romero2014fitnets}, relational knowledge distillation~\cite{park2019relational} and contrastive and adversarial approaches to distillation~\cite{tian2019contrastive}. Our work builds on these great works by proposing a multi-resolution distillation method that aligns features at different feature resolutions between teacher and student, enabling a more efficient and effective knowledge transfer in dense prediction tasks, such as segmentation.
\section{Method}
\label{Met}
\subsection{Overall architecture}
\label{sec:method}
Inspired by the two-branch architectures~\cite{yu2021bisenet, poudel2019fast, hong2021deep,huang2021alignseg,chen2022mobile}, we design a squeeze-enhanced Axial Transformer (SeaFormer) framework. 
As is shown in Figure~\ref{fig:pipeline}, SeaFormer consists of these parts: \textit{shared STEM}, \textit{context branch}, \textit{spatial branch}, \textit{fusion block} and \textit{light segmentation head}.
For a fair comparison, we follow TopFormer~\cite{zhang2022topformer} to design the STEM. 
It consists of one regular convolution with stride of 2 followed by four MobileNet blocks where stride of the first and third block is 2.
The context branch and the spatial branch share the produced feature map,  which allows us to build a fast semantic segmentation model.

\subsubsection{Context branch}
\begin{figure*}
  \centering
  \includegraphics[width=\hsize]{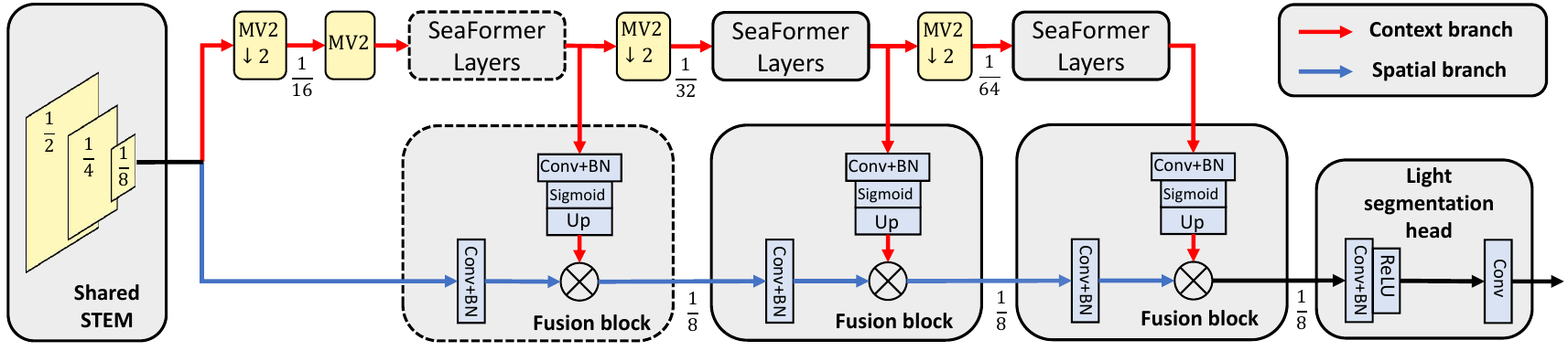}
  \caption{The overall architecture of SeaFormer. It contains shared STEM, context branch (\textbf{red}), spatial branch (\textbf{blue}), fusion block and light segmentation head. 
  \texttt{MV2} block means MobileNetV2 block and \texttt{MV2} {$\downarrow$}2 means MobileNetV2 block with downsampling. 
  SeaFormer layers and fusion block with dash box only exist in SeaFormer-L. The symbol
  $\bigotimes$ denotes element-wise multiplication.
  }
  \label{fig:pipeline}
\end{figure*}

The context branch is designed to capture context-rich information from the feature map $\mathbf{x}_s$.
As illustrated in the red branch of Figure~\ref{fig:pipeline}, the context branch is divided into three stages.
To obtain larger receptive field, we stack SeaFormer layers after applying a MobileNet block to down-sampling and expanding feature dimension. 
Compared with the standard convolution as the down-sampling module, MobileNet block increases the representation capacity of the model while maintaining a lower amount of computation and latency.
For variants except SeaFormer-Large, SeaFormer layers are applied in the last two stages for superior trade-off between accuracy and efficiency. 
For SeaFormer-Large, we insert SeaFormer layers in each stage of context branch.
To achieve a good trade-off between segmentation accuracy and inference speed, we design a squeeze-enhanced Axial attention block (SEA attention) illustrated in the next subsection.

\subsubsection{Spatial branch}
The spatial branch is designed to obtain spatial information in high resolution. 
Identical to the context branch, the spatial branch reuses feature maps $\mathbf{x}_s$. 
However, the feature from the early convolution layers contains rich spatial details but lacks high-level semantic information. 
Consequently, we design a fusion block to fuse the features in the context branch into the spatial branch, bringing high-level semantic information into the low-level spatial information.

\subsubsection{Fusion block}
As depicted in Figure~\ref{fig:pipeline}, high resolution feature maps in the spatial branch are followed by a 1 × 1 convolution and a batch normalization layer to produce a feature to fuse. 
Low resolution feature maps in the context branch are fed into a $1 \times 1$ convolution layer, a batch normalization layer, a sigmoid layer and up-sampled to high resolution to produce semantics weights by bilinear interpolation. 
Then, the semantics weights from context branch are element-wisely multiplied to the high resolution feature from spatial branch.
The fusion block enables low-level spatial features to obtain high-level semantic information.

\subsubsection{Light segmentation head}
The feature after the last fusion block is fed into the proposed segmentation head directly, as demonstrated in Figure~\ref{fig:pipeline}. 
For fast inference purpose, our light segmentation head consists of two convolution layers, which are followed by a batch normalization layer separately and the feature from the first batch normalization layer is fed into an activation layer.

\subsection{Squeeze-enhanced Axial attention}
\label{sec:sea_attn}
The global attention can be expressed as 
\begin{equation}
    \mathbf{y}_o = \sum_{p\in \mathcal{G}(o)} \text{softmax}_p\left(\mathbf{q}_o^\top \mathbf{k}_p \right)\mathbf{v}_p
\end{equation}
where $\mathbf{x}\in \mathbb{R}^{H\times W\times C}$. $\mathbf{q,k,v}$ are linear projection of $\mathbf{x}$, \ie $\mathbf{q}=\mathbf{W}_q\mathbf{x}, \mathbf{k}=\mathbf{W}_k\mathbf{x}, \mathbf{v}=\mathbf{W}_v\mathbf{x}$, where $\mathbf{W}_q, \mathbf{W}_k\in \mathbb{R}^{C_{qk}\times C}, \mathbf{W}_v\in \mathbb{R}^{C_v\times C}$ are learnable weights. $\mathcal{G}(o)$ means all positions on the feature map of location $o=(i, j)$.
When traditional attention module is applied on a feature map of $H\times W\times C$, the time complexity can be $\mathcal{O}(H^2 W^2(C_{qk}+C_v))$, leading to low efficiency and high latency. 

\begin{align}
    \label{equ:nei_attn} \mathbf{y}_o &= \sum_{p\in \mathcal{N}_{m\times m}(o)}\text{softmax}_p\left(\mathbf{q}_o^\top \mathbf{k}_p \right)\mathbf{v}_p
\end{align}
\begin{equation}
\begin{split}
    \label{equ:axial_attn} \mathbf{y}_o = \sum_{p\in \mathcal{N}_{1\times W}(o) }\text{softmax}_p\left(\mathbf{q}_o^\top \mathbf{k}_p \right)\mathbf{v}_p \\
    ~~~+ \sum_{p\in \mathcal{N}_{H\times 1}(o) }\text{softmax}_p\left(\mathbf{q}_o^\top \mathbf{k}_p \right)\mathbf{v}_p
\end{split}
\end{equation}

To improve the efficiency, there are some works~\cite{liu2021swin,huang2019ccnet,ho2019axial} computing self-attention within the local region. 
We show two most representative efficient Transformer in Equation~\ref{equ:nei_attn},~\ref{equ:axial_attn}.
Equation~\ref{equ:nei_attn} is represented by window-based attention~\cite{luong2015effective} successfully reducing the time complexity to $\mathcal{O}(m^2HW(C_{qk}+C_v))=\mathcal{O}(HW)$, where $\mathcal{N}_{m\times m}(o)$ means the neighbour $m\times m$ positions of $o$, but loosing global receptiveness.
The Equation~\ref{equ:axial_attn} is represented by Axial attention~\cite{ho2019axial}, which only reduces the time complexity to $\mathcal{O}((H+W)HW(C_{qk}+C_v))=\mathcal{O}((HW)^{1.5})$, where $\mathcal{N}_{H\times 1}(o)$ means all the positions of the column of $o$; $\mathcal{N}_{1\times W}(o)$ means all the positions of the row of $o$.

According to their drawbacks, we propose the mobile-friendly squeeze-enhanced Axial attention, with a succinct squeeze Axial attention for global semantics extraction and an efficient convolution-based detail enhancement kernel for local details supplement.
\begin{equation}
\begin{split}
\label{equ:hori_ver_squ}
    \mathbf{q}_{(h)} = \frac{1}{W}\left(\mathbf{q}^{\rightarrow(H,C_{qk}, W)}\mathbf{A}_W^{\rightarrow(H,W,1)}\right)^{\rightarrow(H, C_{qk})} \\
    \mathbf{q}_{(v)} = \frac{1}{H}\left(\mathbf{q}^{\rightarrow(W, C_{qk}, H)}\mathbf{A}_H^{\rightarrow(W,H,1)}\right)^{\rightarrow(W, C_{qk})}
    \end{split}
\end{equation}

\subsubsection{Squeeze Axial attention}
\begin{figure}[tb]
  \centering
   \includegraphics[width=0.8\hsize]{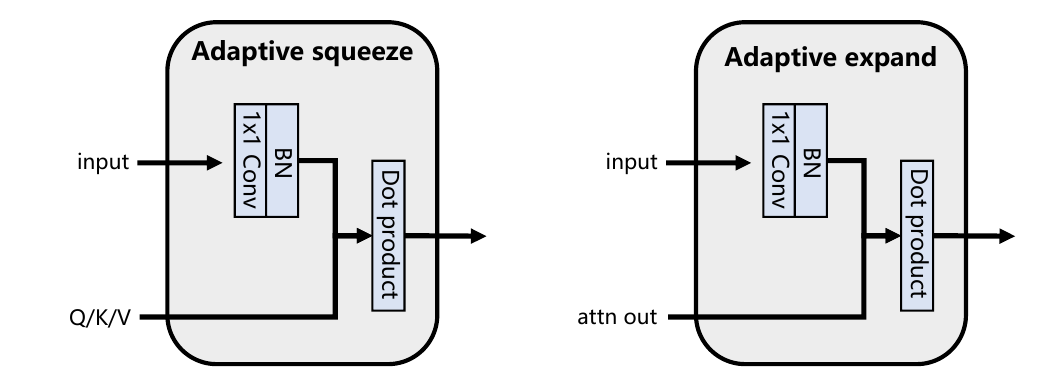}
  \caption{\textit{\textbf{Left}}: the schematic diagram of the proposed adaptive squeezing.
  \textit{\textbf{Right}} is the adaptive expanding operation. \textit{Mat mul} means matrix multiplication. \textit{Attn out} is the output of the multi-head attention.
  }
  \label{fig:squeeze_expand}
\end{figure}

To achieve a more efficient computation and aggregate global information at the same time, we resort to a more radical strategy.
In the same way, $\mathbf{q},\mathbf{k},\mathbf{v}$ are first get from $\mathbf{x}$ with $\mathbf{W}_q^{(s)}, \mathbf{W}_k^{(s)}\in \mathbb{R}^{C_{qk}\times C}, \mathbf{W}_v^{(s)}\in \mathbb{R}^{C_v\times C}$.
According to Equation~\ref{equ:hori_ver_squ}, we first implement \textit{horizontal squeeze} by employing an input adaptive approach, using a learnable mask to map all tokens of query to a single token in each row.
In the same way, the second row of the equation shows the \textit{vertical squeeze} in the vertical direction.
$\mathbf{z}^{\rightarrow(\cdot)}$ means permuting the dimension of tensor $\mathbf{z}$ as given, and $\mathbf{A}$ is a learnable mask that is adaptively adjusted according to the input feature $\mathbf{x}$. It is formed by applying a 1 $\times$ 1 convolution and batch normalization layer on the input feature map \textbf{x}
The adaptive squeeze operation on $\mathbf{q}$ also repeats on $\mathbf{k}$ and $\mathbf{v}$, so we finally get $\mathbf{q}_{(h)}, \mathbf{k}_{(h)}, \mathbf{v}_{(h)}\in\mathbb{R}^{H\times C_{qk}}$, $\mathbf{q}_{(v)}, \mathbf{k}_{(v)}, \mathbf{v}_{(v)}\in\mathbb{R}^{W\times C_{qk}}$.
The squeeze operation reserves the global information to a single axis in an adaptive manner, thus greatly alleviating the following global semantic extraction shown by Equation~\ref{equ:saa}.
\begin{equation}
\begin{split}
\label{equ:saa}
    \mathbf{y}_{(i,j)} = \sum_{p=1}^{H}\text{softmax}_p\left( \mathbf{q}_{(h)i}^\top \mathbf{k}_{(h)p} \right)\mathbf{v}_{(h)p}\\
    + \sum_{p=1}^W\text{softmax}_p\left( \mathbf{q}_{(v)j}^\top \mathbf{k}_{(v)p} \right)\mathbf{v}_{(v)p}
\end{split}
\end{equation}
Each position of the feature map propagates information only on two squeezed axial features.
Similar to adaptive squeezing operation, a 1 $\times$ 1 convolution and batch normalization layer is used to generate an input-adaptive mask for feature restoration. The detail is shown in Figure~\ref{fig:squeeze_expand}. Compared with the pooling operation for squeezing and broadcast for expanding, the adaptive squeezing and expanding operations help the model aggregate spatial information in an input-adaptive way without introducing excessive computational overhead. The empirical study confirms the effectiveness of the approach.
Time complexity for adaptive squeezing $\mathbf{q},\mathbf{k},\mathbf{v}$ is $\mathcal{O}(HW(2C_{qk}+C_v))$, the attention operation takes $\mathcal{O}((H^2+W^2)(C_{qk}+C_v))$ time and the adaptive expanding takes $\mathcal{O}(HW(2C_{qk}+C_v))$ time.
Thus, our squeeze Axial attention successfully reduces time complexity to $\mathcal{O}(HW)$. 

\subsubsection{Squeeze Axial position embedding} Equation~\ref{equ:hori_ver_squ} are, however, not positional-aware, including no positional information of the feature map.
Hence, we propose squeeze Axial position embedding to squeeze Axial attention.
For squeeze Axial attention, we render both $\mathbf{q}_{(h)}$ and $\mathbf{k}_{(h)}$ to be aware of their position in the squeezed axial feature by introducing positional embedding $\mathbf{r}_{(h)}^{q}, \mathbf{r}_{(h)}^{k}\in \mathbb{R}^{H\times C_{qk}}$, which are linearly interpolated from learnable parameters $\mathbf{B}_{(h)}^{q}, \mathbf{B}_{(h)}^{k}\in \mathbb{R}^{L\times C_{qk}}$.
$L$ is a constant.
In the same way, $\mathbf{r}_{(v)}^q, \mathbf{r}_{(v)}^k\in \mathbb{R}^{W\times C_{qk}}$ are applied to $\mathbf{q}_{(v)}, \mathbf{k}_{(v)}$.
Thus, the positional-aware squeeze Axial attention can be expressed as Equation~\ref{equ:saa_pos}.
\begin{figure*}
  \centering
   \includegraphics[width=\hsize]{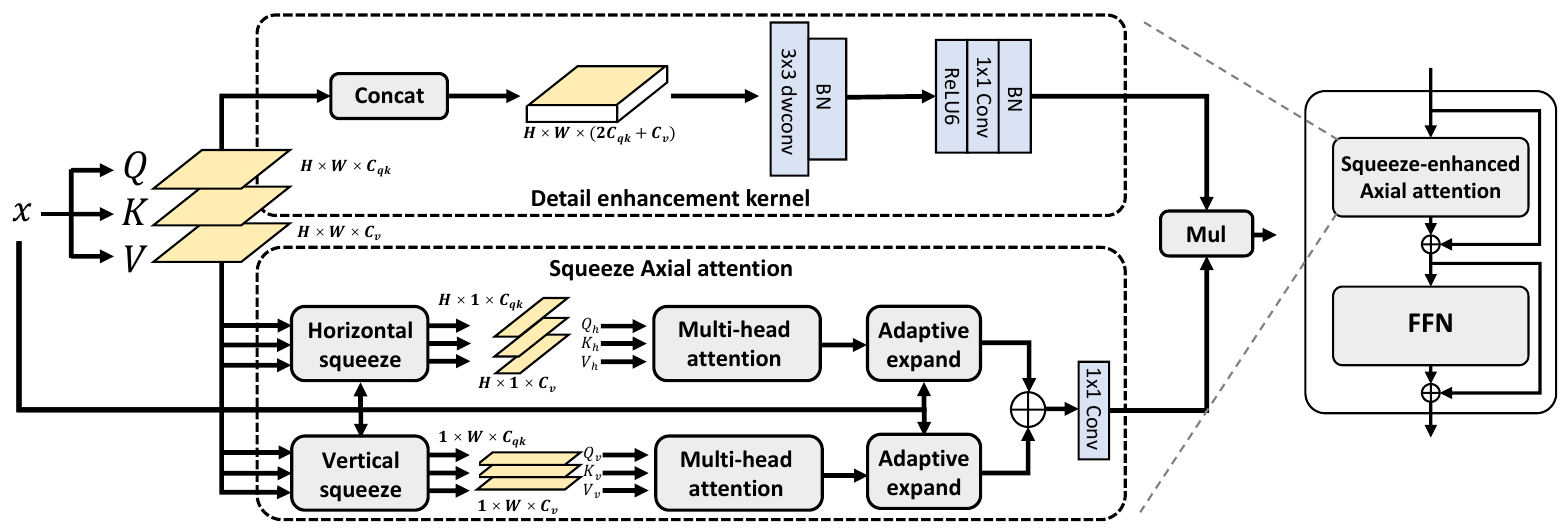}
  \caption{\textit{\textbf{Right}}: the schematic illustration of the proposed squeeze-enhanced Axial Transformer layer including a squeeze-enhanced Axial attention and a Feed-Forward Network (FFN).
  \textit{\textbf{Left}} is the squeeze-enhanced Axial Transformer layer, including detail enhancement kernel and squeeze Axial attention. The symbol $\bigoplus$ indicates an element-wise addition operation. Mul means multiplication.
  }
  \label{fig:transformer}
\end{figure*}
\begin{equation}
\begin{split}
\label{equ:saa_pos}
    \mathbf{y}_{(i,j)} = \sum_{p=1}^{H}\text{softmax}_p\left((\mathbf{q}_{(h)i} + \mathbf{r}_{(h)i}^{q})^\top (\mathbf{k}_{(h)p} + \mathbf{r}_{(h)p}^{k})\right)\mathbf{v}_{(h)p}\\
    + \sum_{p=1}^W\text{softmax}_p\left((\mathbf{q}_{(v)j} + \mathbf{r}_{(v)j}^{q})^\top (\mathbf{k}_{(v)p} + \mathbf{r}_{(v)p}^{k}) \right)\mathbf{v}_{(v)p}
\end{split}
\end{equation}

\subsubsection{Detail enhancement kernel}
The squeeze operation, though extracting global semantic information efficiently, sacrifices the local details. 
Hence an auxiliary convolution-based kernel is applied to enhance the spatial details. 
As is shown in the upper path of Figure~\ref{fig:transformer}, $\mathbf{q},\mathbf{k},\mathbf{v}$ are first get from $\mathbf{x}$ with another $\mathbf{W}_q^{(e)}, \mathbf{W}_k^{(e)}\in \mathbb{R}^{C_{qk}\times C}, \mathbf{W}_v^{(e)}\in \mathbb{R}^{C_v\times C}$ and are concatenated on the channel dimension and then passed to a block made up of 3$\times$3 depth-wise convolution and batch normalization. 
By using a 3$\times$3 convolution, auxiliary local details can be aggregated from 
$\mathbf{q},\mathbf{k},\mathbf{v}$. 
And then a linear projection with activation function and batch normalization is used to squeeze $(2C_{qk}+C_{v})$ dimension to $C$ and generate detail enhancement weights.
Finally, the enhancement feature will be fused with the feature given by squeeze Axial attention. 
Different enhancement modes including element-wise addition and multiplication will be compared in the experiment section.
Time complexity for the 3$\times$3 depth-wise convolution is $\mathcal{O}(3^2HW (2C_{qk}+C_v))$ and the time complexity for the 1$\times$1 convolution is $\mathcal{O}(HWC(2C_{qk}+C_{v}))$. 
Time for other operations like activation can be omitted.

\subsubsection{Complexity analysis}
In our application, we set $C_{qk} = 0.5C_v$ to further reduce computation cost.
The total time complexity of squeeze-enhanced Axial attention is 
\begin{equation}
\begin{split}
    \notag&\mathcal{O}((H^2+W^2)(C_{qk}+C_v) + \mathcal{O}(2HW(2C_{qk} + C_v)) \\
    &+ \mathcal{O}((HWC+9HW)(2C_{qk}+C_{v})) \\
    &=\mathcal{O}((1.5H^2+1.5W^2+4HW)C_v) \\ 
    &+ \mathcal{O}((2HWC+18HW)C_v) \\
    &=\mathcal{O}(HW)
\end{split}
\end{equation}
if we assume $H=W$ and take the channel as constant. SEA attention is linear to the feature map size theoretically.
Moreover, SEA Attention only includes mobile-friendly operations like convolution, pooling, matrix multiplication and so on.

\subsubsection{Architecture details and variants}
SeaFormer backbone contains 6 stages, corresponding to the shared STEM and context branch in Figure~\ref{fig:pipeline} in the main paper. When conducting the image classification experiments, a pooling layer and a linear layer are added at the end of the context branch.

Table~\ref{model_var} details the family of our SeaFormer configurations with varying capacities.
We construct SeaFormer-Tiny, SeaFormer-Small,  SeaFormer-Base and SeaFormer-Large models with different scales via varying the number of SeaFormer layers and the feature dimensions. 
We use input image size of $512 \times 512$ by default. 
For variants except SeaFormer-Large, SeaFormer layers are applied in the last two stages for a superior trade-off between accuracy and efficiency. 
For SeaFormer-Large, we apply the proposed SeaFormer layers in each stage of the context branch.
\begin{figure*}
  \centering
  \includegraphics[width=\hsize]{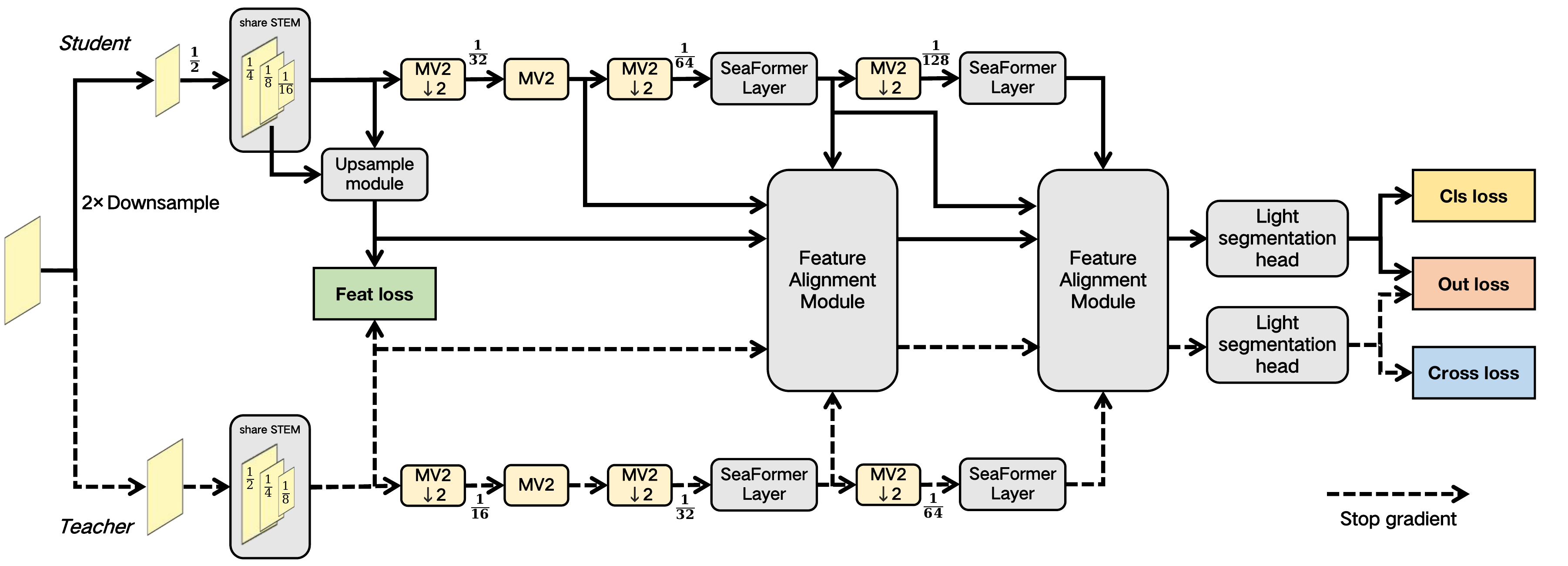}
  \caption{The overall pipeline of multi-resolution distillation based on feature up-sampling. \texttt{MV2(E=4)} denotes MobileNetV2 block with an expansion ratio of 4, and the default kernel size for depth-wise convolution is 5.}
  \label{fig:distill_pipeline}
\end{figure*}
\begin{table*}
\small

  \centering
  \resizebox{1\textwidth}{!}{
  \begin{tabular}{c | c |c  c  c c}
    \hline

    \hline

    \hline
    & Resolution & SeaFormer-Tiny & SeaFormer-Small & SeaFormer-Base & SeaFormer-Large\\
    \hline
    
    \hline
    \hline
    \multirow{2}*{Stage1} & \multirow{2}*{H/2 × W/2} & [Conv, 3, 16, 2] & [Conv, 3, 16, 2] & [Conv, 3, 16,  2]& [Conv, 3, 32, 2] \\
      & & [MB, 3, 1, 16, 1] & [MB, 3, 1, 16, 1] & [MB, 3, 1, 16, 1] & [MB, 3, 3, 32, 1]\\
     \hline
     \multirow{2}*{Stage2} & \multirow{2}*{H/4 × W/4}  
      & [MB, 3, 4, 16, 2] & [MB, 3, 4, 24, 2] & [MB, 3, 4, 32, 2] & [MB, 3, 4, 64, 2]\\
      & & [MB, 3, 3, 16, 1] & [MB, 3, 3, 24, 1] & [MB, 3, 3, 32, 1] & [MB, 3, 4, 64, 1] \\
    \hline
     \multirow{2}*{Stage3} & \multirow{2}*{H/8 × W/8}  
     & [MB, 5, 3, 32, 2] & [MB, 5, 3, 48, 2] & [MB, 5, 3, 64, 2] & [MB, 5, 4, 128, 2] \\
     & & [MB, 5, 3, 32, 1] & [MB, 5, 3, 48, 1] & [MB, 5, 3, 64, 1] & [MB, 5, 4, 128, 1]\\
	\hline
     \multirow{3}*{Stage4} & \multirow{3}*{H/16 × W/16}  
     & [MB, 3, 3, 64, 2] & [MB, 3, 3, 96, 2] & [MB, 3, 3, 128, 2] & [MB, 3, 4, 192, 2]\\
     & & [MB, 3, 3, 64, 1] & [MB, 3, 3, 96, 1] & [MB, 3, 3, 128, 1] & [MB, 3, 4, 192, 1]\\
     & & & & & [Sea, 3, 8] \\
     \hline
     \multirow{2}*{Stage5} & \multirow{2}*{H/32 × W/32}  
     & [MB, 5, 3, 128, 2] & [MB, 5, 4, 160, 2] & [MB, 5, 4, 192, 2] & [MB, 5, 4, 256, 2]\\
     & & [Sea, 2, 4] & [Sea, 3, 6] & [Sea, 4, 8] & [Sea, 3, 8]\\
     \hline
     \multirow{2}*{Stage6} & \multirow{2}*{H/64 × W/64}  
     & [MB, 3, 6, 160, 2] & [MB, 3, 6, 192, 2] & [MB, 3, 6, 256, 2] & [MB, 3, 6, 320, 2]\\
     & & [Sea, 2, 4] & [Sea, 3, 6] & [Sea, 4, 8] & [Sea, 3, 8]\\
\hline

\hline

\hline
  \end{tabular}
  }
  \caption{Architectures for semantic segmentation. [Conv, 3 ,16, 2] denotes regular convolution layer with kernel of 3, output channel of 16 and stride of 2. [MB, 3, 4, 16, 2] means MobileNetV2~\cite{sandler2018mobilenetv2} block with kernel of 3, expansion ratio of 4, output channel of 16 and stride of 2. [Sea, 2, 4] refers to SeaFormer layers with number of layers of 2 and heads of 4.}
  \label{model_var}
\end{table*}

\subsection{Multi-resolution distillation based on feature up-sampling}

In dealing with dense prediction tasks, accurately extracting semantic information and spatial details from images often necessitates the input of high-resolution images. This approach undoubtedly increases the computational burden on the model, especially in resource-constrained environments such as mobile devices or real-time applications, where the demand for high computation becomes even more significant. In recent years, to alleviate this issue, some research~\cite{qi2021multi, hu2022cross} has proposed methods utilizing multi-scale distillation techniques. In these methods, a teacher model that takes high-resolution images as input is used to guide a student model that takes low-resolution images as input to learn. This strategy allows the student model to produce reasonable results even with low-resolution inputs, thereby significantly reducing computational costs and increasing inference speed.

Inspired by the aforementioned methods, this paper proposes an innovative multi-resolution distillation approach based on feature upsampling. Unlike previous work, our method involves aligning the features of the teacher and student models at the same processing stage (even though the student model's input resolution is half that of the teacher model) and particularly emphasizes upsampling at the feature level to achieve alignment.

Specifically, assuming the student model's input resolution is half that of the teacher model, the resolution of the feature maps produced by the student model at any given forward propagation stage will naturally be half that of the corresponding stage in the teacher model. To address this mismatch, this study has designed a feature alignment module. This module employs a lightweight feature upsampling module constructed using MobileNetV2 to upsampling the student model's feature maps to the same resolution as that of the teacher model at the same stage. Subsequently, a feature similarity loss function is used to optimize the similarity between these two features, maximizing their consistency to better aid the student model in mimicking the behavior of the teacher model.

Furthermore, the upscaled features from the student model are not only used for alignment with the teacher model's features but are also integrated into the overall semantic segmentation framework, serving as spatial branch features in the feature fusion process. The specific method of this feature fusion and the involved modules are detailed in Section~\ref{sec:method}. Through this series of steps, the fused features are finally fed into a lightweight segmentation head to complete the semantic segmentation task.

\subsection{Feature alignment module}
\begin{figure}[ht]
  \centering
  \subfloat[The schematic diagram of the feature alignment module.]{%
    \includegraphics[width=0.45\linewidth]{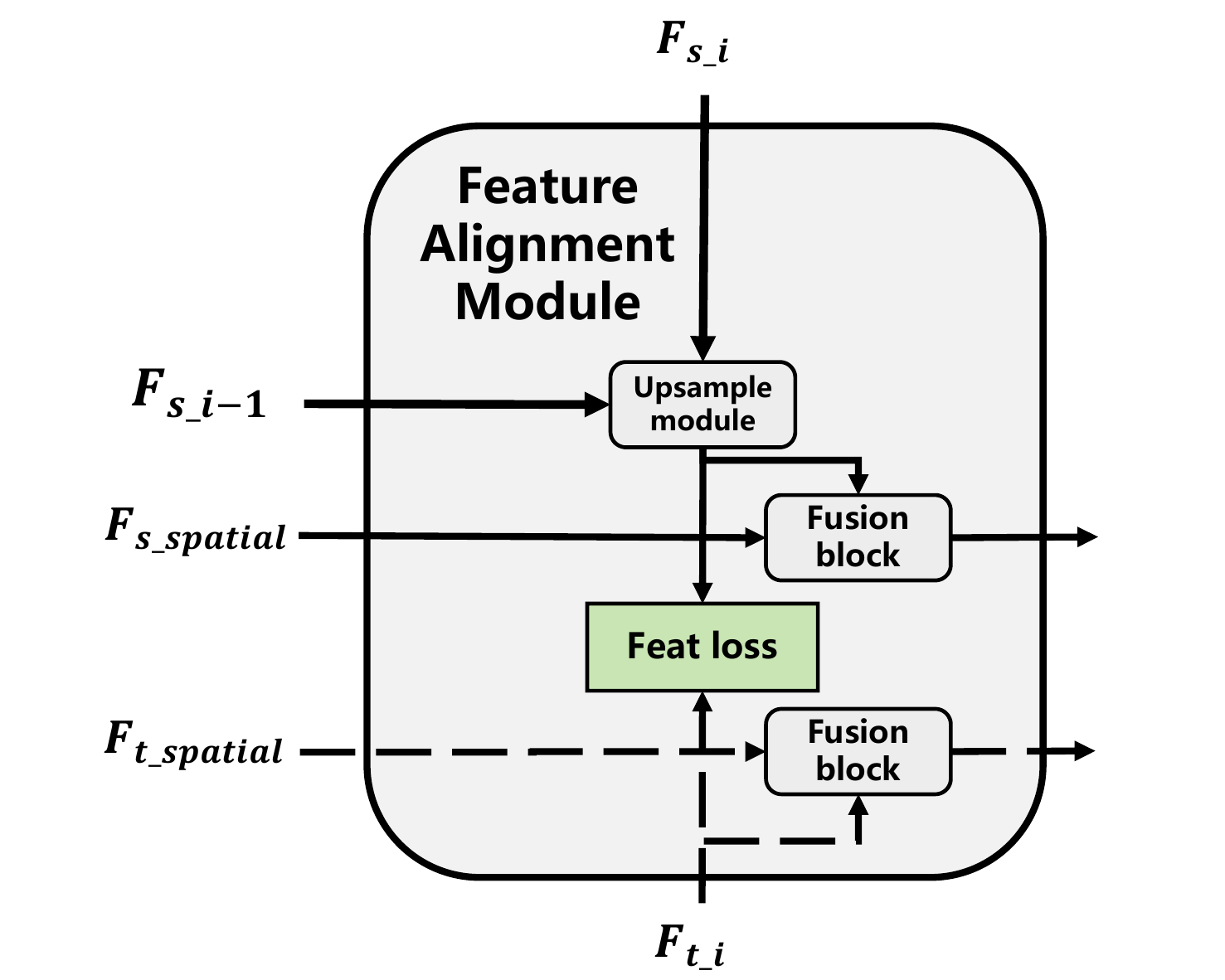}%
    \label{fig:feat_align}%
  }
  \quad 
  \subfloat[The schematic diagram of the upsample module.]{%
    \includegraphics[width=0.45\linewidth]{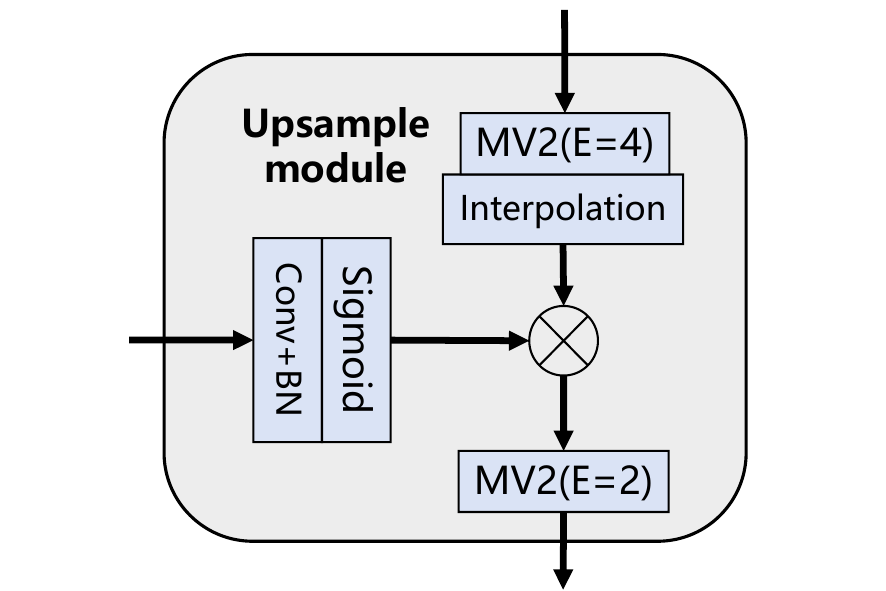}%
    \label{fig:upsample}%
  }
   \caption{\textit{\textbf{Left}}: the schematic illustration of the proposed feature alignment module including an upsample module and a fusion block. \textit{$F_{s\_i}$} means feature map of student model on stage \textit{i}. \textit{$F_{t\_i}$} means feature map of teacher model on stage \textit{i}. \textit{$F_{s\_spatial}$} means feature map of student spatial branch. \textit{$F_{t\_spatial}$} means feature map of teacher spatial branch.
  \textit{\textbf{Right}} is the upsample module.}
  \label{fig:feat_align_upsample}
\end{figure}

Figure~\ref{fig:feat_align} shows the feature alignment module's goal to precisely align the student model's features with the teacher model's, maintaining architectural consistency as described in Section~\ref{sec:method}. This includes using feature fusion to merge spatial and current stage features, and employing the teacher's features as guides to improve the student's feature alignment and predictive accuracy. Initially, the student model's input resolution is halved via average pooling. To address this, a lightweight upsampling module, informed by the detailed matching features of both models, upsamples the student's features for alignment. Cosine similarity between the upscaled and teacher's features, with negative similarity contributing to the loss function, enhances the student's emulation of the teacher's features, ensuring efficient and accurate model performance despite lower input resolution.

\subsubsection{Upsampling module}
Before alignment, the features of the student model are upsampled by applying a lightweight up-sampling module, which ensures that the feature resolution matches the feature resolution of the teacher model at the same stage, facilitating knowledge transfer and improving the performance of the student model. As demonstrated in Figure~\ref{fig:distill_pipeline}, high-resolution feature maps from the previous stage are followed by a convolution, a batch normalization layer, and a sigmoid layer to produce weights to guide the up-sampling of low-resolution features in the current stage. 
Low-resolution feature maps are fed into a  MobileNetV2 block and up-sampled to high resolution.
Then, they are element-wise multiplied with each other, and the up-sampled features are refined with another MobileNetV2 block.

\subsubsection{Loss function}
As indicated in equation~\ref{eqa:distill_loss}, the loss function of the multi-resolution distillation method based on feature up-sampling comprises four components: classification loss, cross-model classification loss, feature similarity loss, and output similarity loss.
\begin{equation}
\begin{split}
\label{eqa:distill_loss}
\mathcal{L}=\mathcal{L}_{cls}+\mathcal{L}_{cross}+\mathcal{L}_{feat}+\mathcal{L}_{out} \\ 
\end{split}
\end{equation}

{\bf Classification loss} $\mathcal{L}_{cls}$ refers to the cross-entropy loss between the output of the student model and the ground truth labels. 

{\bf Cross-model classification loss} $\mathcal{L}_{cross}$ denotes the cross-entropy loss between the output obtained by inputting the up-sampled features of the student model into the segmentation head of the teacher model and the ground truth labels. 

The {\bf feature similarity loss} $\mathcal{L}_{feat}$ measures the negative cosine similarity between the up-sampled features of the student model and the features of the teacher model at the corresponding stage. 

The {\bf output similarity loss} $\mathcal{L}_{out}$ represents the Kullback-Leibler Divergence between the output logits of the student model and the output logits of the teacher model.

The cross-model loss, feature similarity loss, and output similarity loss all contribute to the process of knowledge distillation. Although they differ in terms of the emphasis on the transmitted knowledge from the teacher model to the student model and the degree of relaxation in guiding model parameter updates, they all positively impact the learning of the student model. Our empirical study validates the effectiveness of the aforementioned loss functions.

\begin{table*}
\small
\centering

\begin{tabular}{L{3.6cm} L{1.9cm}| r |r| r |r}
\hline

\hline

\hline
Backbone  & Decoder & Params  &FLOPs & mIoU & Latency\\
\hline

\hline
\hline
MobileNetV2 &  LR-ASPP &  2.2M & 2.8G & 32.0 & 177ms \\
MobileNetV3-Lr &  LR-ASPP &  1.6M & 1.3G & 32.3 & 81ms \\
MobileNetV3-Large &  LR-ASPP &  3.2M & 2.0G & 33.1 & 126ms \\
HRNet-W18-Small  &  HRNetW18S  &  4.0M  & 10.2G  & 33.4  & 639ms \\
TopFormer-T &  Simple Head &  1.4M  & 0.6G  & 33.6  & 43ms   \\
TopFormer-T* &  Simple Head &  1.4M  & 0.6G  & 34.6  & 43ms   \\
PP-MobileSeg-T*&  Seg Head &  1.5M  & 0.7G  & 36.4  & 47ms   \\
\textbf{SeaFormerT}  &  Light Head &  1.7M    & 0.6G   & 35.0      & 40ms\\
\textbf{SeaFormer-T*}  &  Light Head &  1.7M    & 0.6G  & 35.8 & 40ms\\
\textbf{SeaFormer-T++}  &  Light Head &  1.8M    & 0.6G  & \textbf{36.8} & 41ms\\
\textbf{SeaFormer-T++(KD)}  &  Light Head &  2.3M    & 0.3G  & 35.5 & \textbf{22ms}\\
\hline

\hline
\hline
ConvMLP-S &  SemanticFPN &  12.8M & 33.8G & 35.8 & 777ms \\
EfficientNet &  DeepLabV3+  &17.1M & 26.9G & 36.2 & 970ms \\
MobileNetV2 &  Lite-ASPP  & 2.9M & 4.4G & 36.6 & 235ms \\
TopFormer-S  &  Simple Head &  3.1M   & 1.2G   & 36.5  & 74ms   \\
TopFormer-S*  &  Simple Head &  3.1M   & 1.2G   & 37.0  & 74ms   \\
\textbf{SeaFormer-S}  &  Light Head &  4.0M     & 1.1G   & 38.1   & 67ms\\
\textbf{SeaFormer-S*}  &  Light Head &  4.0M    & 1.1G    & 39.4  & 67ms\\
\textbf{SeaFormer-S++}  &  Light Head &  4.1M    & 1.1G    & \textbf{39.7}  & 68ms\\
\textbf{SeaFormer-S++(KD)}  &  Light Head &  5.0M    & 0.5G  & 38.4 & \textbf{33ms}\\
\hline

\hline
\hline
MiT-B0 &  SegFormer &  3.8M & 8.4G & 37.4 & 770ms \\
ResNet18 &  Lite-ASPP  &  12.5M & 19.2G & 37.5 & 648ms \\
ShuffleNetV2-1.5x &  DeepLabV3+ & 16.9M & 15.3G & 37.6 & 960ms\\
MobileNetV2 & DeepLabV3+ & 15.4M  & 25.8G & 38.1 & 1035ms \\
TopFormer-B  &  Simple Head &  5.1M & 1.8G  & 38.3   & 110ms   \\
TopFormer-B*  &  Simple Head &  5.1M & 1.8G  & 39.2   & 110ms   \\
\textbf{SeaFormer-B}  &  Light Head &  8.6M & 1.8G  & 40.2      & 106ms\\
\textbf{SeaFormer-B*}  &  Light Head &  8.6M & 1.8G  & 41.0     & 106ms\\
\textbf{SeaFormer-B++}  &  Light Head &  8.7M & 1.8G  & \textbf{41.4}      & 107ms\\
\textbf{SeaFormer-B++(KD)}  &  Light Head &  10.2M    & 0.9G  & 39.5 & \textbf{55ms}\\
\hline

\hline
\hline
MiT-B1  &  SegFormer &  13.7M & 15.9G  & 41.6   & 1300ms   \\
\textbf{SeaFormer-L}  &  Light Head &  14.0M & 6.5G  & 42.7      & 367ms\\
\textbf{SeaFormer-L*}  &  Light Head &  14.0M & 6.5G  & 43.7      & 367ms\\
\textbf{SeaFormer-L++}  &  Light Head &  14.1M & 6.5G  & \textbf{43.8}      & 369ms\\
\textbf{SeaFormer-L++(KD)}  &  Light Head &  17.1M    & 2.9G  & 42.2 & \textbf{177ms}\\
\hline

\hline

\hline
\end{tabular}
\caption{Results of semantic segmentation on ADE20K \textit{val} set, * indicates training batch size is 32. KD means knowledge distillation.
The latency is measured on a single Qualcomm Snapdragon 865 with input size 512×512, and only an ARM CPU core is used for speed testing. MobileNetV3-Lr means MobileNetV3-Large-reduce. HRNet-W18S means HRNet-W18-Small.
References:
MobileNetV2~\cite{sandler2018mobilenetv2}, 
MobileNetV3~\cite{howard2019searching},
HRNet~\cite{yuan2020object},
TopFormer~\cite{zhang2022topformer},
PP-MobileSeg~\cite{tang2023pp},
ConvMLP~\cite{li2023convmlp},
Semantic FPN~\cite{kirillov2019panoptic},
EfficientNet~\cite{tan2019efficientnet},
DeepLabV3+ and Lite-ASPP~\cite{chen2018encoder},
SegFormer~\cite{xie2021segformer},
ResNet~\cite{he2016deep},
ShuffleNetV2-1.5x~\cite{ma2018shufflenet}.
}
\label{ade20k_table}
\end{table*}
\begin{table*}[tb]
    \centering
  
    \begin{tabular}[b]{L{2.5cm}  L{2.7cm} |R{1.1cm} C{1.3cm} C{1.3cm} R{1.4cm}}
    \hline

    \hline
    
    \hline
    Method     & Backbone     & FLOPs        & mIoU(val) & mIoU(test)  & Latency\\
    \hline

    \hline
    \hline
    FCN & MobileNetV2 & 317G & 61.5 & - &  24190ms\\
    PSPNet  &   MobileNetV2  & 423G & 70.2 & - & 31440ms\\
    SegFormer(h) & MiT-B0 & 17.7G & 71.9 & - & 1586ms\\
    SegFormer(f) & MiT-B0 & 125.5G & 76.2 & - & 11030ms\\
    L-ASPP & MobileNetV2       & 12.6G & 72.7  & - & 887ms \\
    LR-ASPP & MobileNetV3-L      & 9.7G & 72.4  & 72.6 & 660ms  \\
    LR-ASPP & MobileNetV3-S      & 2.9G & 68.4 & 69.4 & 211ms  \\
    SimpleHead(h) & TopFormer-B       &     2.7G      & 70.7   & - & 173ms    \\
    SimpleHead(f) & TopFormer-B       &     11.2G      & 75.0  & 75.0 & 749ms     \\
    \hline

    \hline
    \hline
    Light Head(h) & \textbf{SeaFormerS}       &   2.0G    & 70.7  & 71.0 & 129ms       \\
    Light Head(h) & \textbf{SeaFormerS++}       &   2.0G    & 72.1  & 72.3 & 130ms       \\
    Light Head(f) & \textbf{SeaFormerS}       &   8.0G    & 76.1  & 75.9 & 518ms        \\
    Light Head(f) & \textbf{SeaFormerS++}       &   8.0G    & 77.2  & 76.9 & 521ms       \\
    Light Head(h) & \textbf{SeaFormerB}       &   3.4G    & 72.2  & 72.5 & 205ms        \\
    Light Head(h) & \textbf{SeaFormerB++}       &   3.4G    & 73.5  & 73.4 & 207ms       \\
    Light Head(f) & \textbf{SeaFormerB}       &   13.7G    & 77.7 & 77.5 & 821ms         \\  
    Light Head(f) & \textbf{SeaFormerB++}       &   13.7G    & 78.6  & 78.3 & 825ms       \\
    \hline

    \hline
    
    \hline
  \end{tabular}
\caption{Results on Cityscapes \textit{val} and \textit{test} set. The results on \textit{test} set of some methods are not presented due to the fact that they are not reported in their original papers.}
\label{table_city}
\end{table*}
\section{Experiments}
\label{Exp}
We evaluate our method on semantic segmentation, image classification and object detection tasks.
First, we describe implementation details and compare results with related efficient neural networks.
We then conduct a series of ablation studies to validate the design of SeaFormer.
Each proposed component and important hyper-parameters are examined thoroughly.

\subsection{Experimental setup}
\label{sec:experimental_setup}
\subsubsection{Dataset} We perform semantic segmentation experiments over ADE20K~\cite{zhou2017scene}, CityScapes~\cite{cordts2016cityscapes}, Pascal Context~\cite{mottaghi2014role} and COCO-Stuff~\cite{caesar2018coco}.
The mean of intersection over union (mIoU) is set as the evaluation metric. 
We convert full-precision models to TNN~\cite{contributors2020tnn} and measure latency on an ARM-based device with a single Qualcomm Snapdragon 865 processor.

\paragraph{ADE20K} ADE20K dataset covers 150 categories, containing 25K images that are split into 20K/2K/3K for \textit{Train}, \textit{val} and \textit{test}.

\paragraph{CityScapes} CityScapes is a driving dataset for semantic segmentation. It consists of 5000 fine annotated high-resolution images with 19 categories.

\paragraph{PASCAL Context} dataset has 4998 scene images for training and 5105 images for testing. There are 59 semantic labels and 1 background label.

\paragraph{COCO-Stuff} dataset augments COCO dataset with pixel-level stuff annotations. There are 10000 complex images selected from COCO. The training set and test set consist of 9K and 1K images respectively.

\subsubsection{Implementation details}
We set ImageNet-1K~\cite{deng2009imagenet} pretrained network as the backbone, and training details of ImageNet-1K are illustrated in the last subsection. 
For semantic segmentation, the standard BatchNorm~\cite{ioffe2015batch} layer is replaced by synchronized BatchNorm.
Our implementation is based on public codebase \texttt{mmsegmentation}~\cite{contributors2020mmsegmentation}.
We follow the batch size, training iteration scheduler and data augmentation strategy of TopFormer~\cite{zhang2022topformer} for a fair comparison.

\paragraph{ADE20K} The initial learning rate is 0.0005 for batch size 32 or 0.00025 for batch size 16. The weight decay is 0.01. 
A poly learning rate scheduled with factor 1.0 is adopted.

\paragraph{Cityscapes} The initial learning rate is 0.0003 and the weight decay is 0.01.
The comparison of Cityscapes contains full-resolution and half-resolution.
For the full-resolution version, the training images are randomly scaled and then cropped to the fixed size of 1024 × 1024. 
For the half-resolution version, the training images are resized to 1024 × 512 and randomly scaling, the crop size is 1024 × 512. 

\paragraph{Pascal Context} The initial learning rate is 0.0002 and the weight decay is 0.01. A poly learning rare scheduled with factor 1.0 is used. 

\paragraph{COCO-Stuff} The initial learning rate is 0.0002 and the weight decay is 0.01. A poly learning rare scheduled with factor 1.0 is used. 

During inference, we set the same resize and crop rules as TopFormer to ensure fairness.

\subsection{Comparison with state of the art}

\paragraph{ADE20K}
Table~\ref{ade20k_table} shows the results of SeaFormer and previous efficient backbones on ADE20K \textit{val} set. 
The comparison covers Params (model parameters), FLOPs (floating point operations), Latency and mIoU. 
As shown in Table~\ref{ade20k_table}, SeaFormer outperforms these efficient approaches with comparable or less FLOPs and lower latency. 
Compared with specially designed mobile backbone, TopFormer, which sets global self-attention as its semantics extractor, SeaFormer achieves higher segmentation accuracy with lower latency. 
And the performance of SeaFormer-B++ surpasses MobileNetV3 by a large margin of +8.3\% mIoU with lower latency (-16\%). The results demonstrate that our SeaFormer layers improve the representation capability significantly.

\paragraph{Cityscapes}
From the table~\ref{table_city}, it can be seen that SeaFormer-B++ is 1.3 points better than SeaFormer-B with only a slight increase in latency, showing the benefit of our efficient architecture design with multiple SeaFormer layers embedded in.
It is worth noting that with less computation cost and latency, our SeaFormer-S and SeaFormer-S++ even outperform TopFormer-B. This result further confirms the performance and efficiency of our model when processing high-resolution input images.

\paragraph{Pascal Context}
We compare SeaFormer with the previous approaches on Pascal Context validation set in Table~\ref{pascal_table}. We evaluate the performance over 59 categories and 60 categories (including background). From the results, it can be seen that SeaFormer-S++ is +1.2\% mIoU higher (46.31\% vs.45.08\%) than SeaFormer-S with similar latency. 

\paragraph{COCO-Stuff}
We compare SeaFormer with the previous approaches on COCO-Stuff validation set in Table~\ref{coco-stuff_table}. From the results, it can be seen that SeaFormer-S++ is +1.2\% mIoU higher (34.04 \vs 32.82) than SeaFormer-S with a similar computation cost.
\begin{table*}
\centering
    \begin{tabular}{ll |c | c}
        \hline

        \hline

        \hline
    Backbone     & Decoder     & FLOPs     &mIoU (60/59) \\
        \hline

        \hline
        \hline
    MBV2-s16 &  DeepLabV3+  &  22.24G &  38.59/42.34 \\
    ENet-s16 & DeepLabV3+  & 23.00G & 39.19/43.07 \\
        \hline

        \hline
        \hline
    MBV3-s16 &  LR-ASPP &  2.04G & 35.05/38.02 \\
    TopFormer-T  &  Simple Head &  0.53G  & 36.41/40.39   \\
    SeaFormer-T  &  Light Head &  0.51G   & 37.27/41.49  \\
    \textbf{SeaFormer-T++}  &  Light Head &  0.52G   & \textbf{38.61}/\textbf{42.56}  \\
        \hline

        \hline
        \hline    
    TopFormer-S &  Simple Head &  0.98G   & 39.06/43.68   \\
    SeaFormer-S  & Light Head &  0.98G  & 40.20/45.08   \\
    \textbf{SeaFormer-S++}  & Light Head &  1.00G  & \textbf{41.44}/\textbf{46.31}   \\
        \hline

        \hline
        \hline   
    TopFormer-B & Simple Head &  1.54G   & 41.01/45.28   \\
    SeaFormer-B   & Light Head & 1.57G    & 41.77/45.92   \\
    \textbf{SeaFormer-B++}   & Light Head & 1.60G    & \textbf{42.52}/\textbf{46.40}   \\
        \hline

        \hline

        \hline
  \end{tabular}
\caption{Results on Pascal Context \textit{val} set. We omit the latency as the input resolution is almost the same as that in table 1.}
\label{pascal_table}

\end{table*}
\begin{table*}
\vspace{-0cm}
\centering
    \begin{tabular}{ll |c | c}
        \hline

        \hline
        
        \hline
    Backbone  &  Decoder  & FLOPs     & mIoU  \\
        \hline

        \hline
        \hline
    MBV2-s8 & PSPNet& 52.94G & 30.14  \\
    ENet-s16& DeepLabV3+  & 27.10G & 31.45\\
        \hline

        \hline
        \hline
    MBV3-s16 & LR-ASPP & 2.37G &  25.16 \\
    TopFormer-T  & Simple Head & 0.64G & 28.34   \\
    SeaFormer-T & Light Head & 0.62G & 29.24   \\
   \textbf{ SeaFormer-T++} & Light Head & 0.63 & \textbf{30.76}   \\
        \hline

        \hline
        \hline  
    TopFormer-S  & Simple Head & 1.18G & 30.83  \\
    SeaFormer-S  & Light Head & 1.15G  & 32.82    \\
    \textbf{SeaFormer-S++}  & Light Head & 1.17G  & \textbf{34.04}    \\
        \hline

        \hline
        \hline    
    TopFormer-B  & Simple Head & 1.83G & 33.43\\
    SeaFormer-B  & Light Head &  1.81G  & 34.07 \\
    \textbf{SeaFormer-B++}  & Light Head &  1.84G  & \textbf{35.01} \\
        \hline

        \hline
        
        \hline
  \end{tabular}
\caption{Results on COCO-Stuff \textit{test} set. We omit the latency in this table as the input resolution is the same as that in table 1.}
\label{coco-stuff_table}

\end{table*}

\subsection{Ablation studies}
In this section, we ablate different self-attention implementations and some important design elements in the proposed model, including our squeeze-enhanced Axial attention module (SEA attention) and fusion block on ADE20K dataset.

\paragraph{The influence of components in SEA attention}
We conduct experiments with several configurations, including detail enhancement kernel only, squeeze Axial attention only, and the fusion of both. 
As is shown in Table~\ref{ablate_attn}, only detail enhancement or squeeze Axial attention achieves a relatively poor performance, and enhancing squeeze Axial attention with detail enhancement kernel brings a performance boost with a gain of 2.3\% mIoU on ADE20K. 
The results indicate that enhancing global semantic features from squeeze Axial attention with local details from convolution optimizes the feature extraction capability of Transformer block. 
For enhancement input, there is an apparent performance gap between upconv(x) and conv(x). And we conclude that increasing the channels will boost performance significantly.
Comparing concat[qkv] and upconv(x), which also correspond to w/ or w/o convolution weight sharing between detail enhancement kernel and squeeze Axial attention, we can find that sharing weights makes our model improve inference efficiency with minimal performance loss (35.8 \vs 35.9).
As for enhancement modes, multiplying features from squeeze Axial attention and detail enhancement kernel outperforms add enhancement by +0.4\% mIoU. 
\begin{table*}[htb]
  \centering

      \begin{tabular}{C{1.0cm} C{1.0cm} |C{1.6cm} C{1.2cm} | C{0.9cm} C{0.8cm}  C{1.0cm} C{0.6cm}  C{0.6cm}}
        \hline

        \hline

        \hline
        Enhance  & Attn & Enhance & Enhance  & \multirow{2}{*}{Params}     & \multirow{2}{*}{FLOPs}    & \multirow{2}{*}{Latency}     & \multirow{2}{*}{Top1}     &  \multirow{2}{*}{mIoU} \\
        kernel      &  branch   & input &  mode   &        &        &         &     &   \\
        \hline

        \hline
        \hline
        \ding{52}  &  & - &  -  &  1.3M  & 0.58G  & 38ms &  65.9  & 32.5 \\
          &   \ding{52} & - &  -   & 1.4M & 0.57G & 38ms & 66.3 & 33.5 \\
        \ding{52} & \ding{52} & conv(x) & Mul  & 1.6M     & 0.60G     & 40ms    & 67.2   & 34.9   \\
        \ding{52} & \ding{52} & upconv(x) & Mul  & 1.8M     & 0.62G     & 41ms    & 68.1   & 35.9   \\
        \ding{52} & \ding{52} & concat[qkv] & Mul  & 1.7M     & 0.60G     & 40ms    & 67.9   & 35.8   \\
        \ding{52} & \ding{52} & concat[qkv] & Add  & 1.7M     & 0.60G     & 40ms    & 67.3   & 35.4   \\
        \hline

        \hline

        \hline
    \end{tabular}
    \caption{Ablation studies on components in SEA attention on ImageNet-1K and ADE20K datasets. Enhancement input means the input of detail enhancement kernel. conv(x) means x followed by a point-wise conv. upconv(x) is the same as conv(x) except different channels as upconv(x) is from  $C_{in}$ to $C_q+C_k+C_v$ and conv(x) is from $C_{in}$ to $C_{in}$. concat[qkv] indicates concatenation of Q,K,V.}
    \label{ablate_attn}
\end{table*}
\begin{table*}[htb]
  \centering

      \begin{tabular}{c c  | c c c c c}
        \hline

        \hline

        \hline
        Squeeze method  & Expand method   & {Params}     & {FLOPs}    & {Latency}     & {Top1}     &  {mIoU} \\
        \hline

        \hline
        \hline
        Mean pooling & Broadcast  & 1.7M     & 0.60G    & 40ms    & 67.9  & 35.8   \\
        Max pooling & Broadcast  & 1.7M     & 0.60G     & 40ms    & 67.4   & 35.0   \\
        Adaptive squeeze & Adaptive expand & 1.8M     & 0.61G     & 41ms    & 69.8   & 36.8   \\
        \hline

        \hline

        \hline
    \end{tabular}
    \caption{Ablation studies on squeeze and expand methods in SEA attention on ImageNet-1K and ADE20K datasets.}
    \label{ablate_adasqueeze}
\end{table*}

\paragraph{Comparing different self-attention modules in the Swin Transformer}
\begin{table}
    \centering
    \small
 
    \begin{tabular}[b]{l | c |c |c | c}

    \hline
    Model     & Params(B)     & FLOPs(B)        & mIoU  & Latency\\
    \hline

    Swin	&27.5M	&25.6G	&44.5	&3182ms\\
    CCNet	&41.6M	&37.4G	&43.1	&3460ms\\
    ISSA	&31.8M	&33.3G	&37.4	&2991ms\\
    A2-Nets	&37.2M	&31.1G	&28.9	&2502ms\\
    Axial	&36.2M	&32.5G	&45.3	&3121ms\\
    Local	&27.5M	&25.1G	&34.2	&3059ms\\
    MixFormer	&27.5M	&24.9G	&45.5	&2817ms\\
    ACmix	&27.9M	&26.6G	&45.3	&3712ms\\
    Global	&27.5M	&0.144T	&OOM	&14642ms\\

    \hline

    \textbf{SeaFormer}	&34.0M	&\textbf{24.9G}	&\textbf{46.5} &\textbf{2278ms}\\

    \hline
  \end{tabular}
\caption{Results on ADE20K \textit{val} set based on Swin Transformer architecture. (B) denotes backbone. OOM means CUDA out of memory. References: ISSA~\cite{huang2019interlaced}, A2-Nets~\cite{chen20182}}
\label{table_swin} 
\end{table}
To eliminate the impact of our architecture and demonstrate the effectiveness and generalization ability of SEA attention, we ran experiments on Swin Transformer~\cite{liu2021swin} by replacing window attention in Swin Transformer with different attention blocks.
We set the same training protocol, hyper-parameters, and model architecture configurations as Swin for a fair comparison. 
When replacing window attention with CCAttention (CCNet) or DoubleAttention (A2-Nets), they have much lower FLOPs than SeaFormer and other attention blocks. 
Considering that we may not be able to draw conclusions rigorously, we doubled the number of their Transformer blocks (including MLP).
As ACmix has the same architecture configuration as Swin, we borrow the results from the original paper. 
From Table~\ref{table_swin}, it can be seen that SeaFormer outperforms other attention mechanisms with lower FLOPs and latency.

\paragraph{Comparing different self-attention modules in the SeaFormer}
To verify the effectiveness and efficiency of SEA attention based on our designed pipeline, we experiment with convolution, Global attention, Local attention, Axial attention and three convolution enhanced attention methods including our SEA attention, ACmix and MixFormer. 
The ablation experiments are organized in seven groups.
Since the resolution of computing attention is relatively small, the window size in Local attention, ACmix, and MixFormer is set to 4.
We adjust the channels when applying different attention modules to keep the FLOPs aligned and compare their performance and latency.
The results are illustrated in Table~\ref{table_attn}.

As demonstrated in the table, SEA attention outperforms the counterpart built on other efficient attentions.
Compared with global attention, SEA attention outperforms it by +1.2\% Top1 accuracy on ImageNet-1K and +1.6 mIoU on ADE20K with less FLOPs and lower latency. 
Compared with similar convolution enhanced attention works, ACmix and MixFormer, our SEA attention obtains better results on ImageNet-1K and ADE20K with similar FLOPs but lower latency. 
The results indicate the effectiveness and efficiency of SEA attention module.
\begin{table}

    \centering
    \small

        \begin{tabular}{l | c | c| c| c| c}
        \hline
        
        \hline

        \hline
        Method     & Params   & FLOPs   & Latency     & Top1    &  mIoU  \\
        \hline

        \hline
        \hline
        Conv  &  1.6M  & 0.59G  & 38ms &  66.3  & 32.8 \\
        Local & 1.3M  &  0.60G  & 48ms &  65.9 & 32.8  \\
        Axial & 1.6M  &  0.63G  & 44ms &  66.9 & 33.7  \\
        Global & 1.3M & 0.61G  & 43ms & 66.7 & 34.2 \\
        ACmix &   1.3M  & 0.60G & 54ms & 66.0 & 33.1 \\
        MixFormer &   1.3M & 0.60G  & 50ms & 66.8 & 33.8 \\
        \textbf{SeaFormer} & 1.7M & 0.60G & \textbf{40ms}  & \textbf{67.9}  & \textbf{35.8}  \\
        \hline

        \hline

        \hline
    \end{tabular}
    \caption{Performance of different self-attention modules on our designed pipeline on ImageNet-1K and ADE20K datasets. 
    }
    \label{table_attn}
\vspace{-0.1cm}
\end{table}

\paragraph{Comparing different architecture with the same attention module}
Table~\ref{tab:ablate_arch} provides a comprehensive comparison of different architectures with the same attention module. Our proposed architecture, SeaFormer++, demonstrates a favorable balance between efficiency and accuracy. Notably, SeaFormer++ achieves a latency of 369ms with only 6.5G FLOPs, which is significantly lower than Swin Transformer (2278ms and 24.9G FLOPs) while maintaining competitive mIoU performance (43.8 vs. 46.5). 

Additionally, with the inclusion of knowledge distillation (SeaFormer++ KD), the model’s latency further improves to an impressive 177ms with only 2.9G FLOPs, while still maintaining a mIoU of 42.2. This demonstrates the impact of knowledge distillation in not only reducing computational cost but also preserving performance. Compared to DeepLabV3+ and SegFormer, our SeaFormer++ with KD achieves comparable accuracy (42.2 mIoU vs. 42.7 mIoU for SegFormer) while offering much faster inference times and fewer FLOPs, confirming the efficiency of the overall architecture design and distillation pipeline.
\begin{table*}
\centering
\begin{tabular}{l|c|c|c|c}
\hline
        
\hline
        
\hline
Architecture   & Params & FLOPs & Latency & mIoU \\ 
\hline
        
\hline   
\hline
DeepLabV3+~\cite{chen2018encoder} & 17.1M  & 24.5G& 710ms & 39.3  \\
SegFormer~\cite{xie2021segformer} &  17.1M & 15.2G& 920ms   & 42.7     \\
Swin~\cite{liu2021swin}	&34.0M	&24.9G	 &2278ms &46.5\\
SeaFormer++ &  14.1M & 6.5G    & 369ms & 43.8\\
SeaFormer++(KD)   &  17.1M    & 2.9G  & 177ms& 42.2 \\
\hline
        
\hline
        
\hline
\end{tabular}
\caption{Comparison of different architecture with our proposed SEA attention. }
\label{tab:ablate_arch}
\end{table*}

\paragraph{The influence of squeeze and expand method}
To evaluate different squeeze and expand strategies within the SEA attention framework, we conducted a structured series of ablation studies. These were divided into three groups based on the method used, focusing on maintaining consistent FLOPs, latency, and comparative performance on the ImageNet-1K and ADE20K datasets.

Table~\ref{ablate_adasqueeze} summarizes the outcomes. Notably, the ‘Adaptive squeeze and Adaptive expand' method excelled, achieving 69.8\% Top1 accuracy on ImageNet-1K and 36.8 mIoU on ADE20K. The 1x1 convolutional layers in the adaptive squeeze and expand modules contribute a minor increase in parameter count (+0.1M) and computational cost (+0.01G), which have a negligible impact on latency (+1ms). However, these modifications result in significant performance improvement, demonstrating the effectiveness of our design choices. 

\paragraph{The influence of fusion block design}
We set four fusion methods, including Add directly, Multiply directly, Sigmoid add and Sigmoid multiply. \textbf{X} directly means features from context branch and spatial branch \textbf{X} directly. 
Sigmoid \textbf{X} means feature from context branch goes through a sigmoid layer and \textbf{X} feature from spatial branch.

From Table~\ref{table_fusion} we can see that replacing sigmoid multiply with other fusion methods hurts performance. 
Sigmoid multiply is our optimal fusion block choice.
\begin{table}[tb]
    \centering
        \begin{tabular}{l|c}
        
        \hline
        
        \hline
        
        \hline
        Fusion method  &  mIoU  \\
        \hline
        
        \hline
        \hline
        Add directly&	35.2\\
        Multiply directly&	35.2\\
        Sigmoid add	&34.8\\
        \textbf{Sigmoid multiply}&	\textbf{35.8}\\
        \hline
        
        \hline
        
        \hline
    \end{tabular}
    \caption{Ablation study on fusion method on ADE20K \textit{val} set.}
        \label{table_fusion}
\end{table}
\begin{table}
    \centering

        \begin{tabular}{c | c |c | c |c}
        \hline
        
        \hline

        \hline
        Embed dim  & Params     & FLOPs    & Latency    &  mIoU \\
        \hline
        
        \hline
        \hline
        64,96        & 8.5M          & 1.7G      & 102ms   & 40.3 \\
        128,160       & 8.6M          & 1.8G      & 106ms   &  41.0\\
        192,256       & 8.7M          & 2.0G      & 121ms   &  41.2\\
        \hline
        
        \hline

        \hline
        \end{tabular}
        \vspace{0.34cm} 
        \begin{tabular}{c | c |c | c |c}
        \hline
        
        \hline

        \hline
        Position bias   & Params     & FLOPs    & Latency    &  mIoU \\
        \hline
        
        \hline
        \hline
            \ding{56} & 1.65M       & 0.60G          & 40ms      & 35.6   \\
            \ding{52} & 1.67M       & 0.60G          & 40ms      & 35.8   \\
        \hline
        
        \hline

        \hline
    \end{tabular}
    \caption{Ablation studies on embedding dimensions and position bias. [128, 160] is an optimal embedding dimension in fusion blocks.}
    \label{table_embdim_posbias}
\end{table}

\paragraph{The influence of the width in fusion block}
To study the influence of the width in fusion block, we perform experiments with different embedding dimensions in fusion blocks on SeaFormer-Base, M denotes the channels that spatial branch and context branch features mapping to in two fusion blocks. Results are shown in Table~\ref{table_embdim_posbias}.

\begin{table*}
\centering
\begin{tabular}{l|l|c|c|c|c}
\hline
        
\hline
        
\hline
Model        & Head    & Params & FLOPs & Latency & mIoU \\ 
\hline
        
\hline   
\hline
Topformer-Tiny*        & Simple head      & 1.4M            & 0.58G          & 43ms             & 34.6          \\ 
Topformer-Tiny*        & Light head       & 1.4M            & 0.64G          & 47ms             & 33.1          \\ 
SeaFormer-Tiny*        & Simple head      & 1.7M            & 0.60G          & 42ms             & 35.6          \\ 
SeaFormer-Tiny*        & Light head       & 1.7M            & 0.60G          & \textbf{40ms}             & 35.8          \\ 
\textbf{SeaFormer-Tiny++}        & Light head       & 1.8M            & 0.61G          & 41ms             & \textbf{36.8}          \\ 
\hline
        
\hline  
\hline
Topformer-Small*       & Simple head      & 3.1M            & 1.16G          & 74ms             & 37.0          \\ 
Topformer-Small*       & Light head       & 3.1M            & 1.19G          & 76ms             & 35.3          \\ 
SeaFormer-Small*       & Simple head      & 4.0M            & 1.10G          & 70ms             & 38.8          \\ 
SeaFormer-Small*      & Light head       & 4.0M            & 1.10G          & \textbf{67ms}             & 39.4          \\ 
\textbf{SeaFormer-Small++}       & Light head       & 4.1M            & 1.11G          & 68ms             & \textbf{39.7}          \\ 
\hline
        
\hline  
\hline
Topformer-Base*        & Simple head      & 5.1M            & 1.81G          & 110ms            & 39.2          \\ 
Topformer-Base*        & Light head       & 5.1M            & 1.85G          & 112ms            & 36.6          \\ 
SeaFormer-Base*        & Simple head      & 8.7M            & 1.84G          & 108ms            & 40.6          \\ 
SeaFormer-Base*        & Light head       & 8.6M            & 1.82G          & \textbf{106ms}            & 41.0          \\ 
\textbf{SeaFormer-Base++}        & Light head       & 8.7M            & 1.83G          & 107ms            & \textbf{41.4}          \\ 
\hline
        
\hline
        
\hline
\end{tabular}
\caption{Comparison of different heads and backbones across various models. * indicates training batch size is 32.}
\label{tab:ablate_backbone_head}
\end{table*}

\paragraph{Additional ablation studies}
To justify the effectiveness of the backbone and decoder in SeaFormer++, we have conducted additional experiments comparing different combinations of backbones and heads. The results are shown in Table~\ref{tab:ablate_backbone_head}.

We used the same backbone with different heads, including the Simple head from the most related baseline Topformer~\cite{zhang2022topformer} and the Light head proposed in our work. Compared to the Simple head in SeaFormer-Base, our proposed Light head gains 0.4 mIoU improvement with a decrease of 0.1M parameters, 0.02G FLOPs, and 2ms Latency. Additionally, we compared different backbones (Topformer, SeaFormer, and SeaFormer++) using the same proposed Light head, SeaFormer-Base++ outperforms TopFormer-Base (41.4 \vs 36.6) with lower latency (107ms \vs 112ms).

The results in the table highlight the superior performance of the SeaFormer++ backbone and Light head in terms of both efficiency (latency and FLOPs) and effectiveness (mIoU). SeaFormer++ consistently outperforms Topformer across various configurations, confirming the strength of our backbone and head design.

\subsection{Image classification}
\label{sec:imgnet_training}
We conduct experiments on ImageNet-1K~\cite{deng2009imagenet}, which contains 1.28M training images and 50K validation images from 1,000 classes. We employ an AdamW~\cite{kingma2014adam} optimizer for 600 epochs using a cosine decay learning rate scheduler. A batch size of 1024, an initial learning rate of 0.064, and a weight decay of 2e-5 are used. The results are illustrated in Table~\ref{imagenet_table}. Compared with other efficient approaches, SeaFormer++ achieves a relatively better trade-off between latency and accuracy.

\section{Object detection}
To further demonstrate the generalization ability of our proposed SeaFormer++ backbone on other downstream tasks, we conduct object detection task on COCO dataset. 

We use RetinaNet~\cite{lin2017focal} (one-stage) as the detection framework and follow the standard settings to use SeaFormer++ as backbone to generate e feature pyramid at multiple scales. All models are trained on train2017 split for 12 epochs (1×) from ImageNet pretrained weights.

From the table~\ref{table_coco_obj}, we can observe that our SeaFormer++ achieves superior results on detection task which further demonstrates the strong generalization ability of our method.

\subsection{Latency statistics}

\begin{figure}
  \centering
   \includegraphics[scale=0.6]{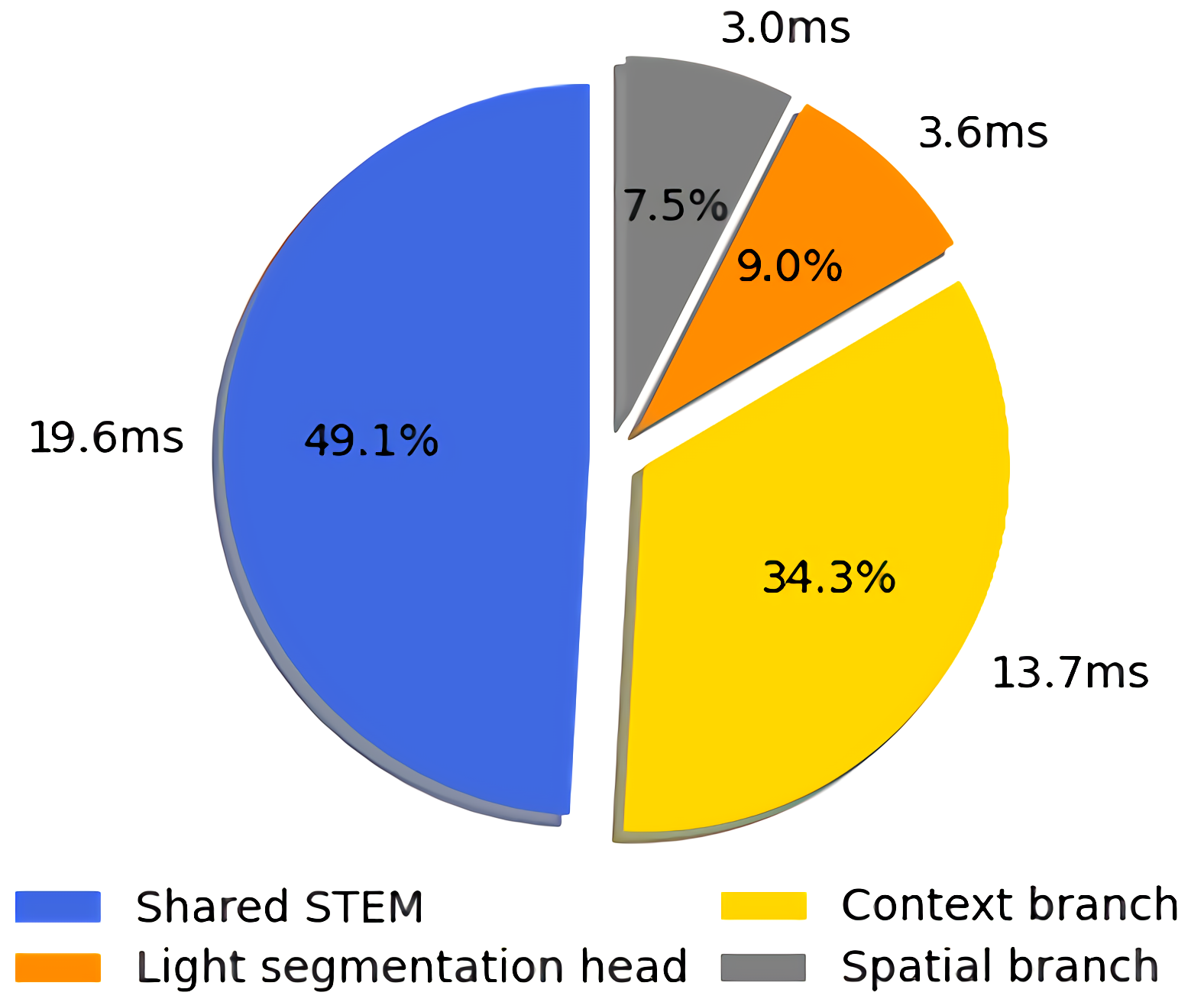}
  \caption{The inference latency of components.}
  \label{lat}
\end{figure}

We make the statistics of the latency of the proposed SeaFormer-Tiny, as shown in Figure~\ref{lat}, the shared STEM takes up about half of the latency of the whole network (49\%). 
The latency of the context branch is about a third of the total latency (34\%), 
whilst the actual latency of the spatial branch is relatively low (8\%) due to sharing early convolution layers with the context branch.
Our light segmentation head (8\%) also contributes to the success of building a light model.

\section{Performance under different precision of the models}
Following TopFormer, we measure the latency in the submission paper on a single Qualcomm Snapdragon 865, and only an ARM CPU core is used for speed testing. No other means of acceleration, e.g., GPU or quantification, is used. We provide a more comprehensive comparison to demonstrate the necessity of our proposed method. We test the latency under different precision of the models.
From the table~\ref{table_diff_precision}, it can be seen that whether it is full precision or half precision the performance of SeaFormer is better than that of TopFormer.

\begin{table*}
\small
\centering

\begin{tabular}{l|c|c|c|c}
\hline

\hline

\hline
Method  & Params  &FLOPs & Top1 & Latency\\
\hline

\hline
\hline
MobileNetV3-Small~\cite{howard2019searching}  & 2.9M & 0.1G & 67.4 & \textbf{5ms} \\
SeaFormer-T  & 1.8M & 0.1G & 67.9 & 7ms \\
\textbf{SeaFormer-T++}  & 1.9M & 0.1G & \textbf{69.8} & 7ms \\
\hline

\hline
\hline
MobileViT-XXS~\cite{mehta2021mobilevit}  & 1.3M & 0.4G & 69.0 & 24ms \\
MobileViTv2-0.50~\cite{mehta2022separable} & 1.4M & 0.5G & 70.2 & 32ms \\
MobileOne-S0*~\cite{vasu2023mobileone} & 2.1M & 0.3G & 71.4 & 14ms \\
MobileNetV2~\cite{sandler2018mobilenetv2} & 3.4M & 0.3G & 72.0 & 17ms \\
Mobile-Former96~\cite{chen2022mobile} & 4.8M & 0.1G & 72.8 & 31ms \\
SeaFormer-S & 4.1M & 0.2G & 73.3 & 12ms \\
\textbf{SeaFormer-S++} & 4.2M & 0.2G & \textbf{74.5} & \textbf{12ms} \\
\hline

\hline
\hline
EdgeViT-XXS~\cite{pan2022edgevits} & 4.1M & 0.6G & 74.4  & 71ms \\
LVT~\cite{yang2022lite} & 5.5M & 0.9G & 74.8 & 97ms \\
MobileViT-XS~\cite{mehta2021mobilevit} & 2.3M & 0.9G & 74.8 & 54ms \\
MobileNetV3-Large~\cite{howard2019searching} & 5.4M & 0.2G & 75.2 & \textbf{16ms} \\
Mobile-Former151~\cite{chen2022mobile} & 7.7M & 0.2G & 75.2 & 42ms \\
MobileViTv2-0.75~\cite{mehta2022separable} & 2.9M & 1.0G & 75.6 & 68ms \\
EfficientFormerV2-S0~\cite{li2023rethinking} & 3.5M & 0.4G & 75.7 & 25ms \\
MobileOne-S1*~\cite{vasu2023mobileone} & 4.8M & 0.8G & 75.9 & 40ms \\
SeaFormer-B & 8.7M & 0.3G & 76.0 & 20ms \\
\textbf{SeaFormer-B++} & 8.8M & 0.3G & \textbf{77.0} & 20ms \\
\hline

\hline
\hline
MobileOne-S2*~\cite{vasu2023mobileone} & 7.8M & 1.3G & 77.4 & 63ms \\
EdgeViT-XS~\cite{pan2022edgevits} & 6.8M & 1.1G & 77.5  & 124ms \\
MobileViTv2-1.00~\cite{mehta2022separable} & 4.9M & 1.8G & 78.1 & 115ms \\
MobileOne-S3*~\cite{vasu2023mobileone} & 10.1M & 1.9G & 78.1 & 91ms \\
AxialFormer~\cite{wang2020axial} & 11.6M & 3.0G & 78.1 & 151ms \\
MobileViT-S~\cite{mehta2021mobilevit} & 5.6M & 1.8G & 78.4 & 88ms \\
EfficientNet-B1~\cite{tan2019efficientnet} & 7.8M & 0.7G & 79.1 & 61ms \\
EfficientFormer-L1~\cite{li2022efficientformer} & 12.3M & 1.3G & 79.2 & 94ms \\
Mobile-Former508~\cite{chen2022mobile}  & 14.8M & 0.5G & 79.3 & 102ms \\
MobileOne-S4*~\cite{vasu2023mobileone} & 14.8M & 3.0G & 79.4 & 143ms \\
SeaFormer-L & 14.0M & 1.2G & 79.9 & 61ms \\ 
\textbf{SeaFormer-L++} & 14.1M & 1.2G & \textbf{80.6} & \textbf{61}ms \\ 
\hline

\hline

\hline
\end{tabular}
\caption{Image classification results on ImageNet-1K \textit{val} set. The FLOPs and latency are measured with input size 224×224, except for MobileViT and MobileViTv2 which are measured with 256×256 according to their original implementations. * indicates re-parameterized variants~\cite{vasu2023mobileone}. The latency is measured on a single Qualcomm Snapdragon 865, and only an ARM CPU core is used for speed testing. No other means of acceleration, e.g., GPU or quantification, is used.}
\label{imagenet_table}
\end{table*}
\begin{figure*}[tbp]
	\centering
	\subfloat[Squeeze Axial attention heatmaps]{\label{heatmap_vis_saa}
	                                        \includegraphics[width=1.2in, height=1.in]{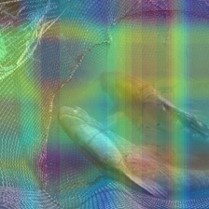}
	                                        \includegraphics[width=1.2in, height=1.in]{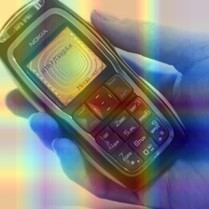}
	                                        \includegraphics[width=1.2in, height=1.in]{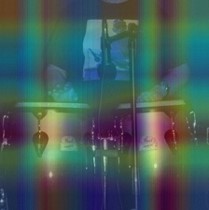}
	                                        \includegraphics[width=1.2in, height=1.in]{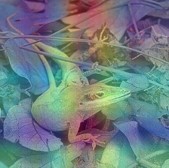}
	                                        }\\
	\subfloat[Squeeze-enhanced Axial attention heatmaps]{\label{heatmap_vis_sea}   
	                                        \includegraphics[width=1.2in, height=1.in]{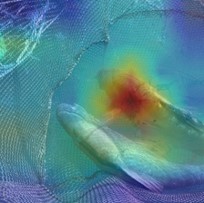}
	                                        \includegraphics[width=1.2in, height=1.in]{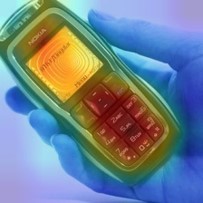}
	                                        \includegraphics[width=1.2in, height=1.in]{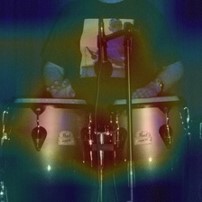}
	                                        \includegraphics[width=1.2in, height=1.in]{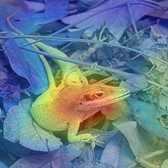}
	                                        }\\	
	\centering
	\caption{The visualization of attention heatmaps from the model consisting of squeeze axial attention without detail enhancement (\textit{first row}) and SeaFormer (\textit{second row}). Heatmaps are produced by averaging channels of the features from the last attention block, normalizing to [0, 255], and up-sampling to the image size.}
	\label{heatmap_vis}
\end{figure*}
\begin{figure*}[tbp]
	\centering
	\subfloat[Ground Truth]{
               \includegraphics[width=1.5in, height=1.in]{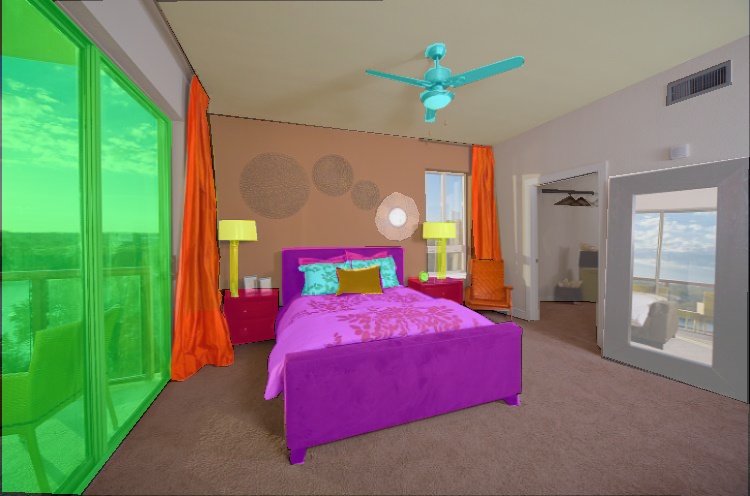} 
               \includegraphics[width=1.5in, height=1.in]{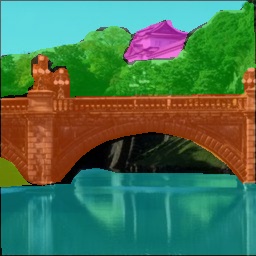} 
               \includegraphics[width=1.5in, height=1.in]{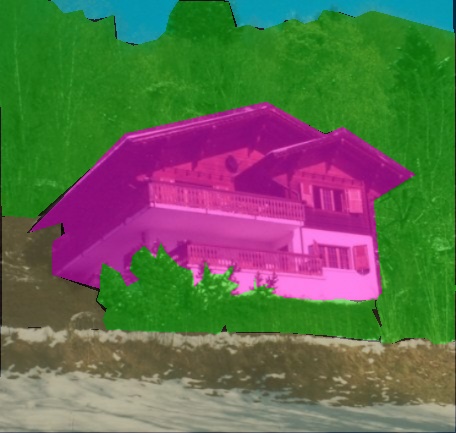} 
	                                        }\\
	\subfloat[TopFormer-B~\cite{zhang2022topformer}]{
                \includegraphics[width=1.5in, height=1.in]{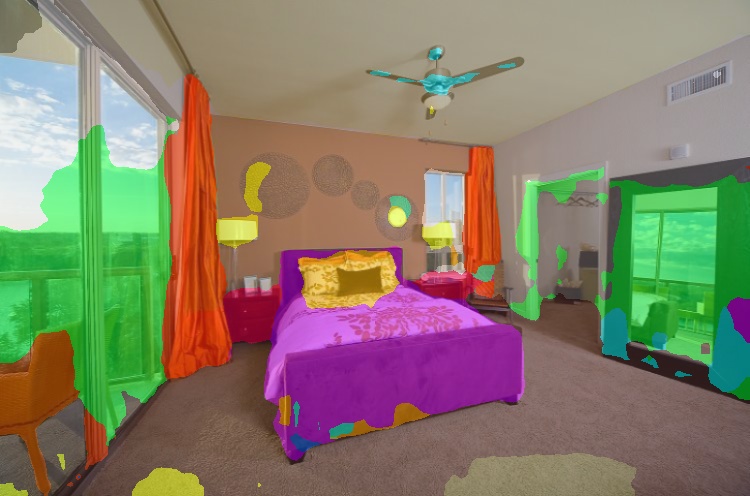} 
                \includegraphics[width=1.5in, height=1.in]{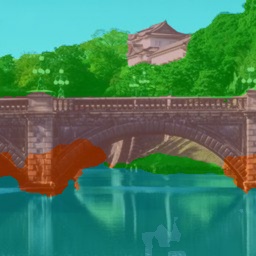} 
                \includegraphics[width=1.5in, height=1.in]{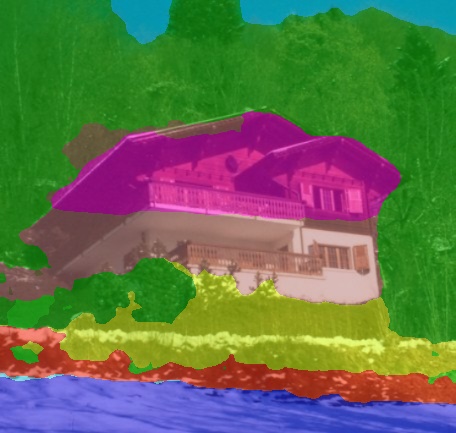} 
	                                        }\\

	\subfloat[SeaFormer-B (Ours)]{
	          \includegraphics[width=1.5in, height=1.in]{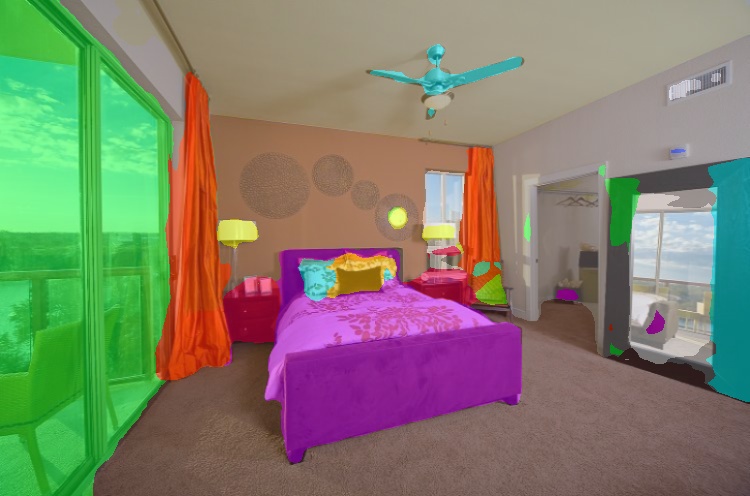} 
	          \includegraphics[width=1.5in, height=1.in]{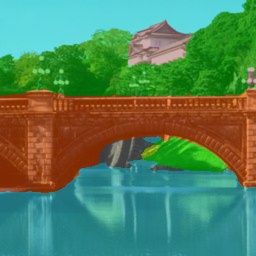} 
	          \includegraphics[width=1.5in, height=1.in]{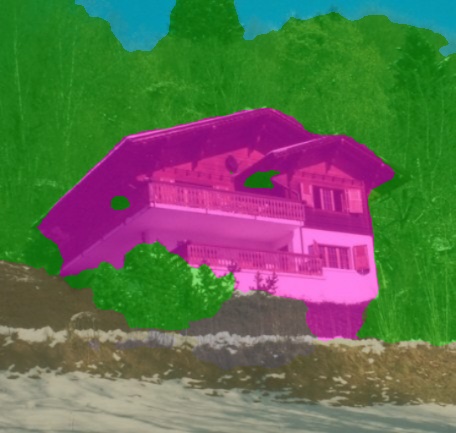} 
	                                        }  \\                                      
	\centering
	\caption{Visualization of prediction results on ADE20K \textit{val} set.}
	\label{vis_results_2}
\end{figure*}

\begin{table}[tb]
    \centering
    \small
    \begin{tabular}[b]{l | c |c |c }
    \hline

    \hline

    \hline
        Backbone & AP  & FLOPs    & Params\\

    \hline
        
    \hline
    \hline
    ShuffleNetv2 & 25.9  & 161G  & 10.4M   \\
    SeaFormer-T &  31.5        & 160G    & 10.9M        \\
    \textbf{SeaFormer-T++} &  \textbf{32.8}         & \textbf{160G}    & 11.0M        \\    
    \hline
          
    \hline
    \hline
    Mobile-Former151   & 34.2   & 161G   & 14.4M\\
    MobileNetV3 & 27.2  &162G   & 12.3M \\

    SeaFormer-S &  34.6           & 161G   & 13.3M         \\
    \textbf{SeaFormer-S++} &  \textbf{35.5}           & \textbf{161G}   & 13.4M         \\  
    \hline
        
    \hline
    \hline
    Mobile-Former214     & 35.8  &\textbf{162G}    & 15.2M\\
    Mobile-Former294     & 36.6   &164G   & 16.1M\\
    SeaFormer-B &  36.7      & 164G   &18.1M            \\
    \textbf{SeaFormer-B++} &  \textbf{37.4}      & 164G   &18.2M            \\
    \hline

    \hline
    \hline
    ResNet50 & 36.5  & 239G & 37.7M \\
    PVT-Tiny& 36.7  & 221G  & 23.0M  \\
    AxialFormer & 37.7  & 210G  & 23.7M  \\
    ConT-M & 37.9  & 217G  & 27.0M   \\
    SeaFormer-L &  39.8      & 185G  &24.0M            \\
    \textbf{SeaFormer-L++} &  \textbf{40.2}       & \textbf{185G}    &24.1M            \\
         \hline

        \hline
        
        \hline
  \end{tabular}
\caption{Results on COCO object detection. References: ShuffleNetv2~\cite{ma2018shufflenet}, Mobile-Former151~\cite{chen2022mobile}, MobileNetV3~\cite{howard2019searching}, ResNet50~\cite{he2016deep}, PVT-Tiny~\cite{wang2021pyramid}, AxialFormer~\cite{wang2020axial}, ConT-M~\cite{yan2021contnet}.}
\label{table_coco_obj}
\end{table}

\begin{figure*}[tbp]
	\centering
	\subfloat[Ground Truth]{
               \includegraphics[width=1.5in, height=1.in]{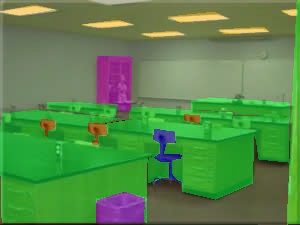} 
               \includegraphics[width=1.5in, height=1.in]{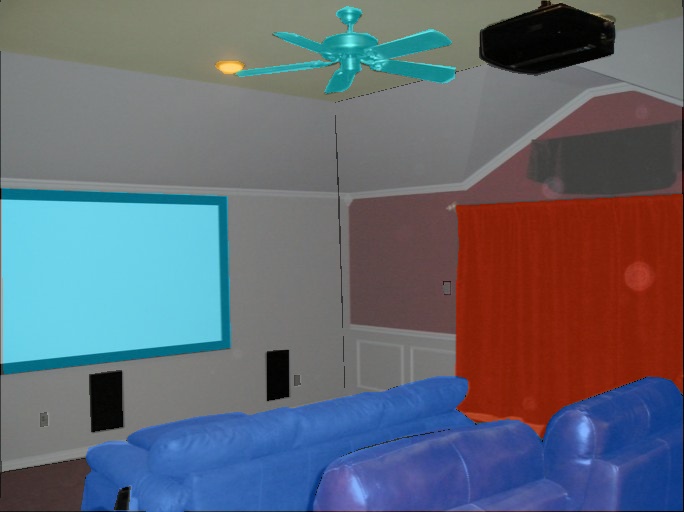} 
               \includegraphics[width=1.5in, height=1.in]{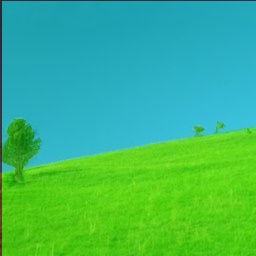} 
	                                        }\\

	\subfloat[SegFormer-B1~\cite{xie2021segformer}]{
	          \includegraphics[width=1.5in, height=1.in]{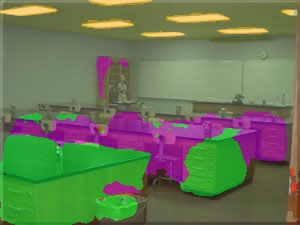} 
	          \includegraphics[width=1.5in, height=1.in]{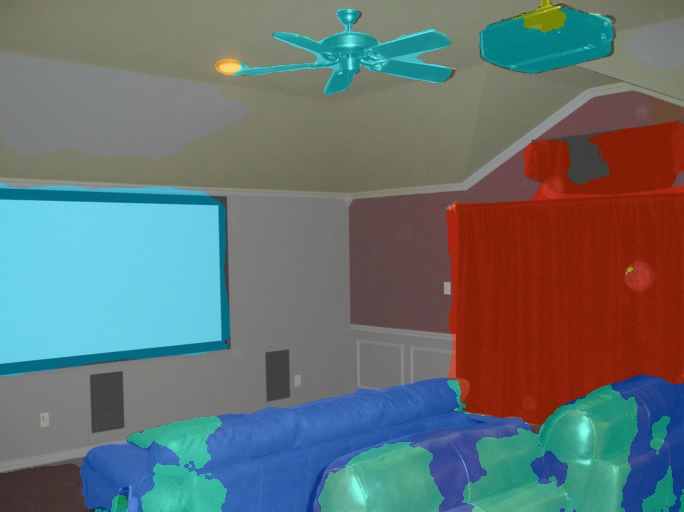} 
	          \includegraphics[width=1.5in, height=1.in]{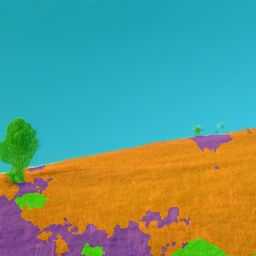} 
	                                        }\\
	\subfloat[SeaFormer-L (Ours)]{
	          \includegraphics[width=1.5in, height=1.in]{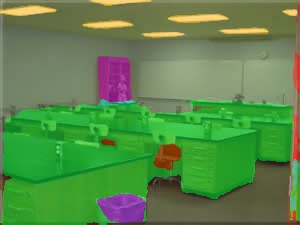} 
	          \includegraphics[width=1.5in, height=1.in]{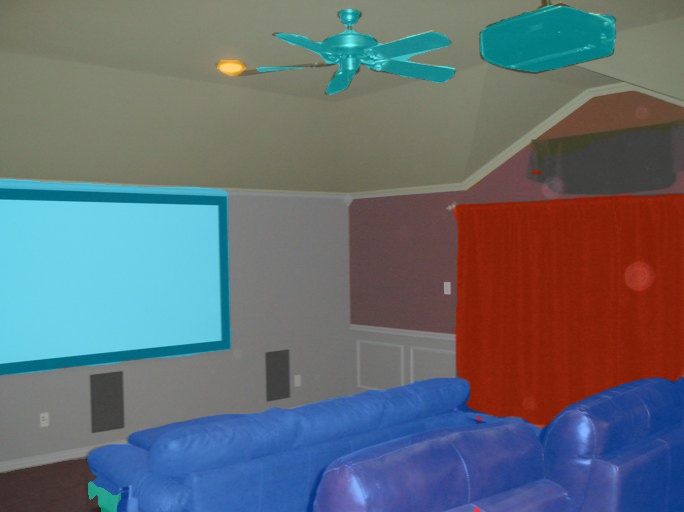} 
	          \includegraphics[width=1.5in, height=1.in]{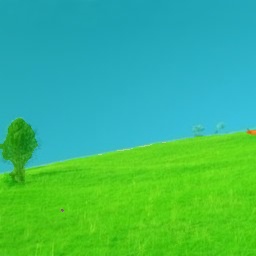} 
	                                        }\\                                        
	\centering
	\caption{Visualization of prediction results on ADE20K \textit{val} set.}
	\label{vis_results}
\end{figure*}
\begin{table}[tb]
    \centering

        \begin{tabular}{l|c|c|c}
        
        \hline
        
        \hline
        
        \hline
        Model  &  mIoU & FP32 & FP16 \\
        \hline
        
        \hline
        \hline
        TopFormer-T	&34.6	&43ms	&23ms \\
        SeaFormer-T	&35.8	&40ms	&22ms\\
        \textbf{SeaFormer-T++}	&\textbf{36.8}	&\textbf{41ms}	&\textbf{23ms}\\
        \hline
        
        \hline
        \hline
        TopFormer-S	&37.0	&74ms	&41ms\\
        SeaFormer-S	&39.4	&67ms	&36ms\\
        \textbf{SeaFormer-S++}	&\textbf{39.7}	&\textbf{68ms}	&\textbf{37ms}\\
        \hline
        
        \hline
        \hline
        TopFormer-B	&39.2	&110ms	&60ms\\
        SeaFormer-B	&41.0	&106ms	&56ms\\
        \textbf{SeaFormer-B++}	&\textbf{41.4}	&\textbf{107ms}	&\textbf{56ms}\\
        \hline
        
        \hline
        \hline
        SeaFormer-L	&43.7	&367ms	&186ms\\
        \textbf{SeaFormer-L++}	&\textbf{43.8}	&\textbf{369ms}	&\textbf{187ms}\\

        \hline
        
        \hline
        
        \hline
    \end{tabular}
    \caption{Performance comparison on ADE20K \textit{val} set under different precision.}
        \label{table_diff_precision}
\end{table}

\section{Visualization}

\subsection{Attention heatmap}
To demonstrate the effectiveness of detail enhancement in our squeeze-enhanced Axial attention (SEA attention),
we ablate our model by removing the detail enhancement. 
We visualize the attention heatmaps of the two models in Figure~\ref{heatmap_vis}.
Without detail enhancement, attention heatmaps from solely SA attention appear to be axial strips while our proposed SEA attention is able to activate the semantic local region accurately, which is particularly significant in the dense prediction task.

\subsection{Prediction results}
We show the qualitative results and compare with the alternatives on the ADE20K validation set from two different perspectives. 
First, we compare with a mobile-friendly rival TopFormer~\cite{zhang2022topformer} with similar FLOPs and latency in Figure~\ref{vis_results_2}. 
Besides, we compare with the Transformer-based counterpart SegFormer-B1~\cite{xie2021segformer} in Figure~\ref{vis_results}.
In particular, our SeaFormer-L has a lower computation cost than the SegFormer-B1.
As shown in both figures, we demonstrate better segmentation results than both the mobile counterpart and Transformer-based approach.

\section{Multi-resolution distillation based on feature up-sampling}
\subsection{Experimental setup} We perform multi-resolution distillation experiments over ADE20K~\cite{zhou2017scene}. mIoU is set as the evaluation metric. 
We convert full-precision models to TNN~\cite{contributors2020tnn} and measure latency on an ARM-based device with a single Qualcomm Snapdragon 865 processor.
We set ADE20K fine-tuned network with original resolution input in section~\ref{Exp} as the teacher model and ImageNet-1K~\cite{deng2009imagenet} pretrained network as the backbone of the student model.
The input resolution of the student model is halved by average pooling.

The standard BatchNorm~\cite{ioffe2015batch} layer is replaced by synchronized BatchNorm.
The implementation of multi-resolution distillation is based on public codebase \texttt{mmsegmentation}~\cite{contributors2020mmsegmentation}.
We follow the batch size, training iteration scheduler and data augmentation strategy of TopFormer~\cite{zhang2022topformer} and section~\ref{Exp} for a fair comparison.
The initial learning rate is 0.0005 for batch size 32 or 0.00025 for batch size 16. The weight decay is 0.01. 
A poly learning rate scheduled with factor 1.0 is adopted.
During inference, we set the same resize and crop rules as TopFormer to ensure fairness.

\subsection{Comparison with state of the art}
Table~\ref{ade20k_table} shows the results of SeaFormer++ (KD) and previous efficient backbones on ADE20K \textit{val} set. 
The comparison covers Params, FLOPs, Latency and mIoU. 
As shown in Table~\ref{ade20k_table}, SeaFormer++ (KD) outperforms these efficient approaches with extremely less FLOPs and lower latency. 

\begin{table*}[ht]
\centering
\begin{tabular}{lcccc}
\hline

\hline

\hline
Upsampling Module & \#Params & FLOPs &  mIoU & Latency \\
\hline
Direct Interpolation & 1.7M & 0.3G & 33.7 & 20ms\\
MobileNetV2-based upsample module & 2.3M & 0.3G & 35.5 & 22ms \\
Standard convolution-based upsample module & 2.7M & 0.4G & 35.7 & 30ms\\
\hline

\hline

\hline
\end{tabular}
\caption{Ablation study results of the upsampling modules.}
\label{tab:upsampling_ablation}
\end{table*}
\begin{table*}[ht]
\centering
\begin{tabular}{cccccc}
\hline

\hline

\hline
Teacher resolution & Cls loss & Out loss & Feat loss & Cross loss &  mIoU \\
\hline
- &   \checkmark & & & &  32.5\\
512x512 &  \checkmark &\checkmark & & &  33.7\\
  512x512 &  \checkmark &\checkmark &\checkmark & &  34.7\\
  512x512 &  \checkmark &\checkmark &\checkmark &\checkmark &  35.5\\
   256x256 &  \checkmark &\checkmark &\checkmark &\checkmark&  32.1\\
\hline

\hline

\hline
\end{tabular}
\caption{Ablation study results of loss function design. The resolution of the student model is 256x256.}
\label{tab:loss_ablation}
\end{table*}
\subsection{Ablation studies}
This study conducts ablation experiments to evaluate various upsampling modules and loss function configurations for reducing computational demands and maintaining performance in visual tasks. It aims to identify efficient upsampling strategies and optimal loss combinations that preserve model accuracy. Experiments were standardized on the ADE20K validation set to ensure fair comparison and result reliability. The research compares the efficacy of techniques like bilinear interpolation, lightweight MobileNetV2-based upsampling, and standard convolutional upsampling, assessed by mIoU and computational impact. Adjusting loss functions, including classification loss, cross-model classification loss, feature similarity loss, and output similarity loss, further analyzes their role in knowledge distillation effectiveness. These experiments offer insights into balancing efficiency and performance in model design, providing a methodology for exploring model enhancements under computational constraints and guiding optimal configurations for accurate visual recognition in practical applications.

\paragraph{Impact of upsample module design}
We compare three different upsampling strategies: direct bilinear interpolation, MobileNetV2-based upsample module, and standard convolution-based upsample module.

In this section, the effects of three upsampling strategies on model performance were compared. The direct interpolation approach, while requiring the least parameters (1.7M) and computational effort (0.3G FLOPs), resulted in the lowest mIoU (33.7\%) and the least latency (20ms), suggesting limited complexity handling. The MobileNetV2-based lightweight upsampling improved mIoU to 35.5\% with a slight latency increase to 22ms, offering a balanced performance enhancement. The main components in the MobileNetV2-based upsampling module are 1x1 convolutional layers. From the table we conclude that 1x1 convolutional layers in the upsampling module contribute a minor increase in parameter count (+0.6M), which has a negligible impact on latency (+2ms) with a significant performance boost, from mIoU 33.7 to mIoU 35.5, demonstrating the effectiveness of our model design. The standard convolution-based module, although yielding the highest mIoU (35.7\%), did so at the cost of increased parameters (2.7M), computation (0.4G FLOPs), and latency (30ms). These findings highlight a trade-off between performance and speed in upsampling choices, with the MobileNetV2-based module providing an optimal balance for dense prediction tasks on resource-constrained devices.

\paragraph{Impact of loss function}
We incrementally add four loss function components—classification loss, output similarity loss, feature similarity loss, and cross-model classification loss—to assess their contribution to model performance.

Through ablation studies, this experiment evaluates the performance impact of different loss functions in semantic segmentation, involving classification loss, output similarity loss, feature similarity loss, and cross-model classification loss, with mIoU as the evaluation metric. Starting from a baseline mIoU of 32.5 with just classification loss, performance sequentially improves with the addition of output and feature similarity losses, highlighting the benefits of aligning student and teacher model outputs and features for accuracy. The highest mIoU of 35.5 is achieved with the inclusion of cross-model classification loss, emphasizing the effectiveness of combining various constraints on the student model. The results underscore the individual and collective contributions of each loss function to semantic segmentation tasks, particularly the significance of model alignment for substantial performance gains, and the synergistic enhancement of model performance through integrated loss function design. To demonstrate the necessity and effectiveness of multi-resolution distillation. We compare with the conventional distillation baseline with a low-resolution teacher. As seen in the table, mIoU of the low-resolution teacher method is 32.1, which is significantly lower than the 35.5 achieved by our multi-resolution distillation method. This highlights the necessity and effectiveness of multi-resolution distillation for improving model performance.

\section{Conclusion}
\label{con}
In this paper, we propose squeeze-enhanced Axial Transformer ({SeaFormer) for mobile semantic segmentation, filling the vacancy of mobile-friendly efficient Transformer.
Moreover, we create a family of backbone architectures of SeaFormer and achieve cost-effectiveness.
The superior performance on the ADE20K, Cityscapes Pascal Context and COCO-Stuff datasets, and the lowest latency demonstrate its effectiveness on the ARM-based mobile device.
Moreover, we employ a feature upsampling-based multi-resolution distillation technique, significantly reducing the inference latency of our framework. This approach enhances model performance at various resolutions, enabling a high-resolution trained teacher model to instruct a low-resolution student model, thus facilitating efficient semantic understanding and prediction on mobile devices with reduced computational demands.
Beyond semantic segmentation, we further apply the proposed SeaFormer architecture to image classification and object detection problems, demonstrating the potential of serving as a versatile mobile-friendly backbone. 

\section{Data Availability Statement}
The datasets generated during and/or analysed during the current study are available in the Imagenet~\cite{deng2009imagenet} (\url{https://www.image-net.org/}), COCO~\cite{caesar2018coco} (\url{https://cocodataset.org}), ADE20K~\cite{zhou2017scene} (\url{https://groups.csail.mit.edu/vision/datasets/ADE20K/}), Cityscapes~\cite{cordts2016cityscapes} (\url{https://www.cityscapes-dataset.com}), Pascal Context~\cite{mottaghi2014role} (\url{https://cs.stanford.edu/~roozbeh/pascal-context/}) and COCO-Stuff~\cite{caesar2018coco} (\url{https://github.com/nightrome/cocostuff?tab=readme-ov-file}) repositories.

\section*{Acknowledgments}
This work was supported in part by National Natural Science Foundation of China (Grant No. 62376060).


\clearpage
\bibliographystyle{spbasic}      
\bibliography{main}   


\end{document}